\documentclass[review, 3p]{elsarticle}

\usepackage{lineno,hyperref}
\modulolinenumbers[5]

\journal{????}%Pattern Recognition

\usepackage{amssymb}
\usepackage[noend]{algorithmic}
\usepackage{amsmath}
\usepackage{graphicx}
\usepackage{multirow}
\usepackage{color}
\usepackage{float}
\usepackage{subfigure}
\usepackage{algorithm}
\usepackage{algorithmic}
\usepackage{pifont}

\usepackage{amsfonts}

\usepackage{amsthm}

\usepackage[utf8]{inputenc}

\theoremstyle{definition}
\newtheorem*{definition*}{Definition:}
\newtheorem{definition}{Definition:}
\newtheorem*{theorem*}{Theorem}
\newtheorem{theorem}{Theorem}
\newtheorem*{proof_sketch}{Proof Sketch}

\usepackage{url}

\begin{document}

\begin{frontmatter}

\title{Better Boosting with Bandits for Online Learning}
%\tnotetext[mytitlenote]{Fully documented templates are available in the elsarticle package on \href{http://www.ctan.org/tex-archive/macros/latex/contrib/elsarticle}{CTAN}.}

%% Group authors per affiliation:
%\author{Elsevier\fnref{AAAAAAAAAAAAAA}}
%\address{Radarweg 29, Amsterdam}
%\fntext[myfootnote]{Since 1880.}

%\fntext[myfootnote]{Since 1880.}

%% or include affiliations in footnotes:
\author[address_UCL]{Nikolaos Nikolaou\corref{mycorrespondingauthor}}
\cortext[mycorrespondingauthor]{Corresponding author}
\author[address_UoE]{Joseph Mellor}
\author[address_NASA]{Nikunj C. Oza}
\author[address_UoM]{Gavin Brown}

\address[address_UCL]{{Department of Physics \& Astronomy, University College London, Gower Street, London, WC1E 6BT, UK}\\{n.nikolaou@ucl.ac.uk}}
\address[address_UoE]{{Usher Institute of Population Health Sciences and Informatics, Old Medical School, Teviot Place, Edinburgh, EH8 9AG}\\{joe.mellor@ed.ac.uk} }
\address[address_NASA]{{NASA Ames Research Center, Moffett Field, CA 94035, USA}\\{nikunj.c.oza@nasa.gov}}
\address[address_UoM]{{School of Computer Science, University of Manchester, Kilburn Building, Oxford Road, Manchester, M13 9PL, UK}\\{gavin.brown@manchester.ac.uk}}

\begin{abstract}
Probability estimates generated by boosting ensembles are poorly calibrated because of the margin maximization nature of the algorithm. The outputs of the ensemble need to be properly calibrated before they can be used as probability estimates. In this work, we demonstrate that online boosting is also prone to producing distorted probability estimates. In batch learning, calibration is achieved by reserving part of the training data for training the calibrator function. In the online setting, a decision needs to be made on each round: shall the new example(s) be used to update the parameters of the ensemble or those of the calibrator. We proceed to resolve this decision with the aid of bandit optimization algorithms. We demonstrate superior performance to uncalibrated and naively-calibrated on-line boosting ensembles in terms of probability estimation. Our proposed mechanism can be easily adapted to other tasks(e.g. cost-sensitive classification) and is robust to the choice of hyperparameters of both the calibrator and the ensemble.
\end{abstract}


\end{frontmatter}

%\linenumbers

\section{Introduction}

%TODO: add applications(\cite{busa2011ranking} included calibration, yet the experiments were done in a purely offline fashion)

AdaBoost~\cite{FreundSchapire1997} is an extensively studied ensemble learning method, with connections to multiple theoretical frameworks like margin theory~\cite{Schapire1998}, game theory~\cite{Freund1999GT}, functional gradient descent~\cite{Mason1999}, additive logistic regression~\cite{Friedman2000}, probabilistic modelling~\cite{Edakunni2011}, to name but a few. AdaBoost classifiers have been very successful as evidenced by extensive experimental comparisons spanning multiple datasets, such as the ones conducted in~\cite{caruana2006empirical, Delgado2014}, and applications like face detection in phone cameras~\cite{ViolaJones2002} and the Yahoo search engine for ranking webpages~\cite{CossockZhang2008}. The success of AdaBoost is further evidenced in numerous machine learning competitions; indicatively, more than half of the winning Kaggle entries have used gradient boosting\footnote{https://github.com/dmlc/xgboost/tree/master/demo\#machine-learning-challenge-winning-solutions (incomplete list)}.%it was a boosting-based ranker that won Yahoo `Learning to Rank Challenge'\cite{  while

Yet despite its success in classification, ranking and regression tasks, the probability estimates generated by AdaBoost have been known to be poorly \emph{calibrated}, i.e. they deviate from \emph{empirical class probabilities}~\cite{Friedman2000, NiculescuCaruana2005, Mease2006, Mease2008}. When the goal is to generate probability estimates rather than just to classify or rank examples, or when it is to solve a cost-sensitive classification problem, the performance of AdaBoost suffers.

Previous work in \emph{batch learning} --tasks in which all data is available and can be processed at once-- has shown that applying some form of calibration to the scores generated by AdaBoost considerably improves performance, in both probability estimation~\cite{NiculescuCaruana2005} and cost-sensitive classification~\cite{Nikolaou2016} tasks. In batch learning scenarios calibration is achieved by \emph{reserving part of the training data} to train a \emph{calibrator function} --usually a \emph{logistic sigmoid} or an \emph{isotonic regression}-- to map uncalibrated raw scores to probability estimates that \emph{maximize the likelihood of the model}. This is done to avoid \emph{overfitting} by training both the ensemble and the calibrator on the same datapoints. 

\emph{Online learning} deals with scenarios where data arrive \emph{sequentially} --either one datapoint at a time or in \emph{minibatches}-- predictions are required as soon as the new datapoints become available and the learner must update its parameters using \emph{only} the previous datapoint (or minibatch). This way the learner can \emph{adapt to changing} --even \emph{adversarial}-- environments. Online learning is also a preferable option when dealing with very large amounts of data, when batch learning becomes expensive or even infeasible (due to computational limitations or slow convergence of the \emph{generalization error})~\cite{wilson2003, bousquet2008, bengio2012}. Despite the increasing relevance of online learning given the growth of \emph{streaming} and \emph{big data} applications and despite the success of AdaBoost as a classifier and ranker, the probability estimation quality of online boosting ensembles has not yet been studied in the literature. 

In this paper we will demonstrate that online boosting ensembles~\cite{oza2005} also produce uncalibrated probability estimates under the most common \emph{scoring functions}, i.e. ways of generating probability estimates. This motivates the need for calibrating the probability estimates of online boosting ensembles. 

%We will see that the same theoretical properties satisfied by AdaBoost in the batch learning setting are satisfied by Oza \& Russell's algorithm in the online setting.%In this paper we will reformulate their algorithm thus making it easier to examine theoretically through the same framework used in \cite{Nikolaou2016} to demonstrate that AdaBoost produces uncalibrated probability estimates

However, calibration is less straightforward in the online setting. On each round we need to \emph{decide} whether the new example(s) will be used to update the parameters of the ensemble or those of the calibrator. A \emph{naive approach} is to use \emph{a fixed policy of calibrating on every} $N_c$ \emph{rounds}. But how do we set this \emph{hyperparameter} $N_c$? Different combinations of \emph{problem} (data, objective), \emph{ensemble} (base learner, ensemble size, scoring function used) and \emph{calibrator} (calibration function, optimization method used) would call for different values of $N_c$. 

In this work we propose resolving this decision with the aid of \emph{bandit optimization algorithms}~\cite{Lai1985,bergemann2006,Kuleshov2010}. Bandit algorithms allow us to choose among a set of actions (here: `\emph{train on new minibatch}' or `\emph{calibrate on new minibatch}') balancing \emph{exploitation} and \emph{exploration} in \emph{stochastic} or \emph{adversarial}, \emph{stationary} or \emph{non-stationary} settings. They do this by \emph{sampling the distribution of rewards of each of the actions} (here: the \emph{increase in log-likelihood of the model following each action}) and maintaining a model of said distribution that is updated upon each feedback. The action to be taken in the next round is then chosen based on the models of the reward distributions. The different bandit algorithms used --how they model the reward distribution and how they decide which action to take next-- are discussed in Section~\ref{sec:bandits}.

Our bandit-based approach --more specifically UCB1-based policies \cite{Auer2002,garivier43,KaufmannCG12}, and Thompson Sampling \cite{thompson1933,agrawal2011,ChapelleL11}, especially in its discounted-rewards version-- shows \emph{superior performance} to uncalibrated and naively-calibrated (i.e. employing fixed policies of calibrating on every $N_c$ rounds) boosting ensembles in probability estimation. This approach is \emph{very easy to adapt to new objectives} (e.g. cost-sensitive learning tasks). All we need to do is change the reward function to the appropriate one for the task at hand (e.g. decrease in classification risk after an action). Moreover, the method is \emph{very flexible and robust to the ensemble hyperparameter choices}, as it will learn an appropriate policy for alternating between the two actions guided by the corresponding rewards of the two actions.

\section{Background}
This work focusses on binary learning tasks. The examples are considered to be of the form $({\bf{x}}_i, y_i)$, where ${\bf{x}}_i$ is the feature vector of the $i$-th example and $y_i\in \{ -1, 1\}$ is its class label. Extension to the multiclass case is often handled by breaking down the problem into multiple binary ones, so our analysis and its main results can carry over to the multiclass case. We consider the online setting where examples are presented to the learner in $M$ minibatches\footnote{The scenario where the examples are arriving one at a time ($b=1$) is merely a common special case.} of size $b$. On the $n$-th iteration the learner performs the following steps:

\begin{enumerate}
\item Receive new examples ${\bf{x}}_i$, $\forall {\bf{x}}_i \in minibatch_n$
\item Predict the label $\hat{y_i}$ and/or\\the probability estimate $\hat{p}(y_i=1|{\bf{x}}_i)$, $\forall i \in minibatch_n$
\item Get true labels $y_i = f({\bf{x}}_i)$, $\forall {\bf{x}}_i \in minibatch_n$, where $f$ is the labelling function
\item Update learner parameters accordingly
\end{enumerate}

The steps above are intentionally left general enough to describe all learning components encountered in the paper. Our goal is to study the quality of the probability estimates generated by online boosting ensembles and strategies for improving it. Online boosting ensembles consist of multiple base learners, themselves also trained in an online fashion and --as we will see-- the techniques used for improving the probability estimates (both the calibrator and the reward models of the bandits) are also learners trained in an online fashion. All follow the same general approach defined above: they maintain a model with a fixed number of parameters (i.e. memory and computational complexity are constant w.r.t the number of examples seen so far), which they update every time the labels of a new minibatch become available.

%In the next subsection we will discuss the most widely used online counterpart boosting algorithm due to Oza \& Russell~\cite{Oza}. 

\subsection{AdaBoost and Online Boosting}
\label{ssec:boosting}
Adaptive Boosting (AdaBoost)~\cite{FreundSchapire1997} is a batch learning algorithm that constructs an ensemble sequentially across multiple rounds. On each round, a new component is added to the model. The principle behind it is to convert a weak learner --a hypothesis whose predictions are marginally more accurate than random guessing-- into a strong one --one of error arbitrarily close to the irreducible Bayes error rate. To achieve this, it focuses on each round on correcting the mistakes of the previous model. This can be done either by \emph{reweighting} or by \emph{resampling} the dataset on each round, putting more emphasis on examples misclassified in the previous round and less on examples correctly classified in it. Each weak learner is assigned a confidence coefficient based on its predictive performance. Predictions are given by a weighted majority vote among the weak learners, the weight of each learner's vote being its confidence coefficient.

%\footnote{Of course there exist extensions \cite{Grabner} and alternative approaches \cite{AAA} to Oza \& Russell. But for the purposes of extending our findings from batch boosting to online boosting, the latter will be sufficient.}

The most popular algorithm for online boosting is the one proposed by Oza~\cite{oza2005} --henceforth \emph{OnlineBoost}. The pseudocode is given in Algorithm~\ref{alg:OnlineBoost}. The ensemble consists of a fixed number $T$ of components (weak learners). As in batch AdaBoost, the idea is to increase the weight assigned to examples that have been misclassified by previous models and decrease the weight assigned to examples that have been classified correctly. In OnlineBoost, each example is presented to each weak learner sequentially. If the $t$-th weak learner misclassifies an example, the example's weight for the purpose of updating the parameters of the $(t+1)$-th weak learner will increase. Conversely, if the $t$-th weak learner classifies an example correctly, the example's weight for the purpose of updating the $(t+1)$-th weak learner will decrease.

The (expected) weight of the current example is captured by the quantity $\lambda$, which is used as the parameter of the \emph{Poisson} distribution from which the `effective weight' $k$ is drawn. A weight of $\lambda=1$ corresponds to no particular emphasis (be it positive or negative) paid to the current example\footnote{Here the \emph{`resampling'} version of the algorithm is shown, and a weight $k$ corresponds to training the weak learner $k$ times on the current datapoint. If the weak learner accepts weighted instances, we can use a \emph{reweighting} approach, i.e. train the classifier \emph{once} on the current datapoint but assign it a weight of $k$. Moreover, since $k \sim Poisson(\lambda)$, we have that $\lambda = \mathbb{E}[k]$, so $\lambda$ could be used instead of $k$ when the reweighting approach is taken, leading to the same updates, in expectation.}. 

The base learner, \emph{OnlineLearnAlg()}, is called for updating the parameters of the $t$-th weak learner on the $i$-th example. Finally, note that $\lambda^{sc}_t$ is the sum of weights corresponding to correctly classified examples so far by the $t$-th weak learner. Conversely, $\lambda^{sw}_t$ is the sum of weights corresponding to examples misclassified by the $t$-th weak learner so far. 

\begin{algorithm}[htb]
   \caption{OnlineBoost}
   \label{alg:OnlineBoost}
\begin{algorithmic}
\STATE {\bfseries Input}: Number of weak learners $T$, training examples $\{(\mathbf{x}_i, y_i)|i =1, \dots, N\}$ presented one at a time
\STATE  {\bfseries For each $i$ do}:
\STATE  {} \ \ \ \ \ Set example weight $\lambda = 1$
\STATE  \ \ \ \ \ {\bfseries For each $t \in \{1, 2, \dots, T\}$ do}:
\STATE {} \ \ \ \ \ \ \ \ \ \ Set $k$ according to $Poisson(\lambda)$
\STATE {} \ \ \ \ \ \ \ \ \ \ {\bfseries Do $k$ times}:
\STATE {} \ \ \ \ \ \ \ \ \ \ \ \ \ \ \ $h_t \leftarrow OnlineLearnAlg(h_t,  (\mathbf{x}_i, y_i))$
\STATE {} \ \ \ \ \ \ \ \ \ \ {\bfseries If $h_t(\mathbf{x}_i)= y_i$}:
\STATE {} \ \ \ \ \ \ \ \ \ \ \ \ \ \ \ $\lambda^{sc}_t \leftarrow \lambda^{sc}_t +\lambda$
\STATE {} \ \ \ \ \ \ \ \ \ \ \ \ \ \ \ $\epsilon_t = \frac{\lambda^{sw}_t}{\lambda^{sw}_t + \lambda^{sc}_t}$
\STATE {} \ \ \ \ \ \ \ \ \ \ \ \ \ \ \ $\lambda \leftarrow \lambda \times \frac{1}{2(1-\epsilon_t)}$
\STATE {} \ \ \ \ \ \ \ \ \ \ {\bfseries Else}:
\STATE {} \ \ \ \ \ \ \ \ \ \ \ \ \ \ \ $\lambda^{sw}_t \leftarrow \lambda^{sw}_t +\lambda$
\STATE {} \ \ \ \ \ \ \ \ \ \ \ \ \ \ \ $\epsilon_t = \frac{\lambda^{sw}_t}{\lambda^{sw}_t + \lambda^{sc}_t}$
\STATE {} \ \ \ \ \ \ \ \ \ \ \ \ \ \ \ $\lambda \leftarrow  \lambda \times \frac{1}{2\epsilon_t}$
\\\hrulefill
\STATE {\bfseries Prediction}:  On example $(\mathbf{x}, y)$, predict $H(\mathbf{x}) = sign \Big[ \sum_{t=1}^{T}{h_t(\mathbf{x})\log{\frac{1-\epsilon_t}{\epsilon_t}}} \Big]$
\end{algorithmic}
\end{algorithm}

In AdaBoost, an example's weight is adjusted according to the base model's performance on the entire training set. Instead, in OnlineBoost, the weight adjustment is based on the performance of a base model only on the examples presented so far. This is of course something intrinsic to online learning. It also means that the sequence of parameter updates will depend on the order in which the examples are presented. Oza~\cite{oza2005} showed that if a lossless\footnote{Following the terminology of~\cite{oza2005}, we use the term \emph{lossless online learner} to describe one whose output model for a given training set is identical to that of the corresponding batch learner (e.g. online Naive Bayes).} online base learner is used, OnlineBoost converges to the same model as an AdaBoost ensemble of the same size trained on the same dataset, as the number of training examples $N\to\infty$. 

To simplify the subsequent discussion we denote the confidence weight of the $t$-th weak learner with 
\begin{equation}
\beta_t = \log{\frac{1-\epsilon_t}{\epsilon_t}}
\end{equation}
and the ensemble output --the quantity whose sign will equal the final predicted class $H(\mathbf{x})$-- with
\begin{equation}
F(\mathbf{x}) = \sum_{t=1}^{T}{\beta_t h_t(\mathbf{x})}. 
\end{equation}

%The model maintained by OnlineBoost is of fixed size w.r.t. the number of datapoints seen so far. It consists of the weak learners $h_t$ and their corresponding voting weights $\log{\frac{1-\epsilon_t}{\epsilon_t}}$. 

\subsection{Probability Calibration}
In many applications it is desirable to estimate the probability of a given example belonging to each class. Quantifying the uncertainty about our predictions allows us to capture the reliability of a classification, to combine predictions from different sources or to make cost-sensitive decisions e.g. by using \emph{Bayesian Decision Theory} principles as done in~\cite{Nikolaou2016}.

However, it is not always straightforward to obtain probability estimates from the outputs of a classifier. Most classifiers allow their output to be treated as a \emph{score} for each test example ${\bf{x}}$ that indicates `\emph{how positive}' ${\bf{x}}$ is deemed. The act of converting \emph{raw scores} to actual probability estimates is called \emph{calibration}.

Denoting with $N$ the total number of examples, $N_{s}$ the number of examples with score $s\in \mathcal{S}$, were $\mathcal{S}$ is finite, and $N_{+,s}$ the number of positives with score $s$, Zadrozny \& Elkan \cite{ZadroznyElkan2002} give the following definition:
\begin{definition}{\bf Calibrated classifier}  
A classifier is said to be calibrated if the empirical probability of an example with score $s(\bf{x}) = s$ belonging to the positive class, $N_{+,s(\bf{x})}/N_{s(\bf{x})}$, tends to the score value $s$, as $N \to \infty, ~\forall s$.
\end{definition}

In practice, most classifiers generate uncalibrated scores, each learner distorts them in its own way according to its inductive bias. We can improve probability estimation by taking measures to correct these distortions, i.e. by mapping the scores generated by the classifier to probability estimates, that maximize the likelihood of the data --or some other measure of quality of the probability estimation. The two most common approaches for doing this are \emph{logistic calibration} (also known as \emph{Platt scaling})~\cite{Platt1999} and \emph{isotonic regression}~\cite{Robertson1988}. 

Logistic calibration finds a sigmoid mapping $s({\bf{x}}) \mapsto \hat{p}(y=1|{\bf{x}})$. Isotonic regression is non-parametric and more general as it can be used to calibrate scores which exhibit any form of monotonic distortion. It needs more data to avoid overfitting and is less straightforward to adapt to the online setting.\\
A common measure used for evaluating probabilistic predictions, is the \emph{logarithmic loss}\footnote{Often referred to as `\emph{cross entropy}', `\emph{logistic loss}' or simply `\emph{log-loss}'. A common alternative to the log-loss for assessing probability estimates is the \emph{Brier score}, i.e. the mean squared error of the probability estimates of the sample of size $N'$ in question, $BS = \frac{1}{N'}\sum_i(y_i -\hat{p}(y_i=1|{\bf{x}}_i))^2$, denoting positive and negative labels with $y_i=1$ and $y_i=0$, respectively. Both belong to the infinite family of measures known as \emph{scoring rules}, for more information on which, we direct the reader to~\cite{winkler1969scoring, kull2015novel}.}, which is the \emph{negative log-likelihood of the true labels given a probabilistic classifier’s predictions}. Denoting positive examples' labels with $y_i=1$ and negative examples' labels with $y_i=0$, the log-loss over some set of examples is given by
\begin{equation}
\label{eq:logloss}
\mathcal{L}= -\sum_i(y_i\log\hat{p}(y_i=1|{\bf{x}}_i) + (1 - y_i)\log(1 - \hat{p}(y_i=1|{\bf{x}}_i))) \in \mathbb{R}.
\end{equation}

Another way to assess the calibration of a classifier's probability estimates is to use \emph{reliability curves}~\cite{degroot1983comparison}. These are plots of probability estimates versus empirical probabilities. For perfectly calibrated predictions, the curve should be as close as possible to the diagonal as per the definition above. In this paper we will be mainly using the running average of the log-loss across all predictions so far to capture the progress of the quality of the probability estimates generated by each ensemble. We will only occasionally present reliability curves constructed across all predictions, for illustrative purposes.%They have the advantage of showing us in which regions and how the probability estimates deviate from empirical probabilities.

%This would correspond to a situation in which among N instances for which a classifier predicts probability p for class A, the ratio of instances which actually belong to class A is approx. p (for any p and sufficiently large N)

\subsection{The multi-armed bandit problem}
\label{sec:bandits}
We will now introduce the basic principles of \emph{bandit optimization} and the specific techniques we will use in the paper. In Section \ref{sec:bandits_for_cal} we will adapt these techniques to automate the process of calibrating probability estimates for OnlineBoost.

The \emph{multi-armed bandit problem} is a simple model for sequential decision making\cite{Lai1985,bergemann2006,Kuleshov2010}. The name stems from the one-armed bandit machines found in casinos. When the machine's lever or `\emph{arm}' is pulled, a cash reward is received with some probability. In the multi-armed bandit problem we imagine an \emph{agent} confronted with many such machines, all with differing distributions for the rewards that the agent might receive by playing them. The agent wishes to \emph{cumulatively maximise their reward} and can do this by pulling the arm with highest reward in expectation.
However, the \emph{reward distributions} are unknown to the agent, and so must be \emph{learned}. There is an \emph{exploration-exploitation tradeoff} for the agent in this scenario. The agent must balance exploiting the knowledge they have by pulling what they believe is the best arm and exploring the arms to increase the \emph{confidence} in this knowledge. There is a large literature on bandits which considers varying assumptions about the reward distributions and the number of arms. 

More formally, a general class of bandit problems is described by a set of \emph{arms} $\mathcal{A}$ with an associated set of \emph{reward distributions} $\nu_{a,n}(\theta_{a, n})$ for $a \in \mathcal{A}$, denoting by $\theta_{a, n}$ the corresponding \emph{distribution parameters}. The agent interacts with the bandit in a series of rounds. At each \emph{round} $n$, the agent chooses an action $a_n \in \mathcal{A}$ and then receives the \emph{reward} $X_n(a) \sim \nu_{a,n}(\theta_{a, n})$. 
We denote the \emph{expected reward at round $n$} of an arm as $\mu_n(a) = \mathbb{E} \left[ \nu_{a, n}(\theta_{a, n}) \right]$ and so the \emph{largest expected reward at time} $n$ is given by $\mu_n(*) = \max_a \mu_n(a)$. The \emph{regret} of a decision $a_n$ is given by $r_n = \mu_n(*) - \mu_n(a_n)$. The \emph{cumulative regret} after $M$ rounds will be, $R_M = \sum_{n=1}^M r_n$. The goal of the agent is to \emph{minimise the expected cumulative regret}, $\mathbb{E}\left[R_M\right]$.

In the standard problem, the reward distributions are assumed to be \emph{stationary} and \emph{Bernoulli} such that $\nu_{a,n}(\theta_{a, n}) = \nu_{a,n'}(\theta_{a, n'}) = \operatorname{Bernoulli}\left( \theta_{a}\right)$, $\forall n, n'$ and a \emph{finite number of arms} \cite{Lai1985,agrawal2012}.
There are many variations of the problem, ranging from \emph{bounded rewards} \cite{Auer2002}, \emph{adversarial rewards} \cite{Auer2003}, \emph{non-stationary rewards}\cite{DBLP:conf/aldt/Viappiani13, DBLP:conf/aistats/MellorS13, Granmo2010, Garivier2008}, and \emph{infinite number of arms} \cite{Kleinberg2008, Slivkins11, Krause2009}. The bandit problem can be extended in many other ways; for example there is much work on \emph{contextual bandits} \cite{Li2011,lloyd2013, May2012}, i.e. policies that take a context, e.g. an example's feature vector $\bf{x}$ into account when deciding the next arm to pull.

In this work, we will assume that the rewards are \emph{stochastic} (i.e. non-adversarial) but the reward distributions are \emph{not} necessarily stationary. The non-stationarity will be handled by using \emph{discounted rewards} as will be explained in Section \ref{sec:bandits_for_cal}.  We will now discuss the specific bandit strategies used in this paper.

\subsubsection{Policies}
\paragraph{\bf{Thompson Sampling}} \cite{thompson1933} is a popular strategy due to having both theoretical justification\cite{agrawal2011,Kaufmann2012,agrawal2012,DBLP:conf/nips/KordaKM13} and strong empirical performance\cite{ChapelleL11}. The strategy assumes a given family of arm reward distributions $\nu$. A distribution is used to model $P(\theta_a | H_n)$ where $H_n$ is the \emph{history of past actions and associated rewards up to round} $n$. After receiving a reward for a given action $a$ the posteriors can be updated via Bayes rule. The agent chooses an arm by first drawing a sample $\hat{\theta_a}(n) \sim P(\theta_{a} | H_n)$ from each arm $a$. The agent chooses to pull arm $a_n = \max_a \mathbb{E} \left[ \nu ( \hat{\theta_{a}}(n) ) \right] $. This is the arm with the highest mean reward conditioned on the sampled parameters $\hat{\theta_a}(n)$. 

%For the case of Bernoulli rewards, since $\nu$ is Bernoulli the parameter $\theta_a$ is the probability of a unit reward. Due to conjugacy between the Bernoulli and Beta distributions $P(\theta_a | H_t)$ is modelled as a Beta distribution with parameters $\alpha_a(t)$ and $\beta_a(t)$. 
For the case of \emph{Gaussian rewards with known variance} $\sigma^2$, since $\nu$ is Gaussian the unknown parameter $\theta_a = \mu_a$ is the mean reward of the arm. Due to the \emph{self-conjugacy of the Gaussian distribution}, $P(\theta_a | H_n)$ is also modelled as a Gaussian distribution with parameters $\hat{\mu_a}(n)$ and $\hat{\sigma^2_a}(n)$, derived in closed-form. The policy, henceforth `\emph{Gaussian Thompson Sampling}', is shown in Algorithm~\ref{alg:normts}.
\begin{algorithm}%[h!]
\caption{Gaussian Thompson Sampling (with known variance $\sigma^2$))}
\begin{algorithmic}%[1]
%\scriptsize
\STATE Let $\hat{\mu_{a}}(1)=0$,
\STATE \hspace{1.6em} $\hat{\sigma^2_{a}}(1)=1$ \hspace{8em} for $a \in \mathcal{A}.$
\FOR{ $n=1,\dots, M$ } 
\STATE Sample $\hat{\theta}_a(n) \sim \mathcal{N}(\hat{\mu_{a}}(n),\hat{\sigma}^2_{a}(n))$, for $a \in \mathcal{A}$.
\STATE Pull arm $a_n = \arg\!\max_{a} \hat{\theta}_a(n)$
\STATE Let $\hat{\mu}_{a_n}(n+1) = \frac{\hat{\mu}_{a_n}(n)\sigma^2 + X_n(a_n)\hat{\sigma}^2_{a_n}(n) }{\sigma^2 + \hat{\sigma}^2_{a_n}(n)}$
\STATE \hspace{1.6em} $\hat{\sigma}^2_{a_n}(n+1) = \frac{\hat{\sigma}^2_{a_n}(n)\sigma^2}{\hat{\sigma}^2_{a_n}(n) + \sigma^2}$
\STATE Let $\hat{\mu_{a}}(n+1) = \hat{\mu_{a}}(n)$ 
\STATE \hspace{1.6em} $\hat{\sigma}^2_{a}(n+1) = \hat{\sigma}^2_{a}(n)$ \hspace{8em} for $a \in \mathcal{A} \setminus \{ a_n \} $.
\ENDFOR
%\STATE
%\STATE
%\STATE \textbf{Final Decision:}
%\STATE
%\STATE Let \(\Omega(T)\) be $\arg\!\max_i \hat{X}_{i,T}$
%
\end{algorithmic}
\label{alg:normts}
\end{algorithm}

\paragraph{\bf{Upper Confidence Bound (UCB) policies}} refer to a particular class of bandit policy \cite{Auer2002,garivier43,KaufmannCG12}. As the name suggests such policies manage exploration through the use of \emph{upper confidence bounds on the estimates of mean arm rewards}. In this way UCB policies follow a principle of \emph{optimism in the face of uncertainty}. A UCB policy starts by pulling each arm once. After this, for each arm the number of pulls of the arm $k_a(n)$ and an estimate of the mean $ \hat{\mu}_{a}(n) = \frac{1}{k_a(n)} \sum_{j=1}^n X_{j}(a) I\left( a_{j} = a\right)$ is maintained.
This is then combined with a \emph{padding function} $c(k_a(n), t)$ to give a upper confidence bound for the arm of $U_a(n) = \hat{\mu}_a(n)  + c(k_a(n), n)$. The agent chooses to pull arm $a_n = \max_a U_a(n)$.
As an example of a padding function, the one used by \emph{UCB1} \cite{Auer2002} is
\begin{equation*}
c(k_a(n), n) = \sqrt{\frac{2 \ln n}{ k_a(n)}}.
\end{equation*}
Padding functions have been further improved with policies such as KL-UCB\cite{garivier43}. In the same paper an improved version of UCB1 was also introduced (see Proposition 4 of \cite{garivier43}), to which we will henceforth refer as `\emph{UCB1-Improved}' .

\section{Probability Estimates under Online Boosting Ensembles}
\label{sec:prob_est}
It is straightforward to adapt the two most common \emph{scoring functions}, i.e. ways of producing probability estimates for batch AdaBoost ensembles, to the OnlineBoost case. The first choice, is to use the \emph{weighed fraction of base learners voting for the positive class}~\cite{NiculescuCaruana2005}.
\begin{equation}
\label{eq:scores}
s({\bf{x}})= \frac{\sum_{t: h_{t}({\bf{x}})=1 } \beta_t h_{t}({\bf{x}})}{\sum_{t = 1}^{T}\beta_t h_{t}({\bf{x}})} \in [0, 1].
\end{equation}
Another choice, motivated by the view of Boosting as an additive logistic regression procedure~\cite{Friedman2000} is\footnote{Eq. (\ref{eq:scores2}) differs from the one given in~\cite{Friedman2000} by a factor of $2$ that multiplies $F$ in the latter. This is simply because the formulation of OnlineBoost of Algorithm~\ref{alg:OnlineBoost} uses $\beta_t = \log\frac{1-\epsilon_t}{\epsilon_t}$ as the confidence weight of the $t$-th weak learner, while~\cite{Friedman2000} uses $\beta_t = \frac{1}{2}\log\frac{1-\epsilon_t}{\epsilon_t}$. Both forms result in the same weight updates after normalizing the latter and to equivalent predictions $H({\bf{x}})$.}
\begin{equation}
\label{eq:scores2}
s'({\bf{x}}) = \frac{1}{1+e^{-F({\bf{x}})}} \in [0, 1].
\end{equation}
Both scores of the form of Eq.(\ref{eq:scores}) and of the form of Eq.(\ref{eq:scores2}) tend to values close to $0$ and $1$. For the case of the former, this behaviour is connected to the \emph{margin maximization} properties of boosting. The (normalized) \emph{hypothesis (a.k.a. voting) margin} of a training example $({\bf{x}},y)$ under the ensemble $F$ is defined as
\begin{equation}
\label{eq:margin}
margin({\bf{x}},y) = \frac{y F({\bf{x}})}{\sum_{t = 1}^{T}\beta_t} \in [-1, 1].
\end{equation}
It is a combined measure of confidence and correctness of the classification of the example under $F$. Its sign encodes whether the example was correctly classified (positive) or misclassified (negative), while the magnitude of the margin measures the \emph{confidence of the final hypothesis}. AdaBoost \emph{greedily maximizes the margins of the training examples}~\cite{Friedman2000}, promoting correct classifications for which the ensemble is highly confident. In fact, this margin maximizing behaviour of boosting algorithms has been connected to their nice generalization properties as classifiers~\cite{Schapire1998}. Theorem \ref{thm:thm1}, given below, shows that OnlineBoost --like AdaBoost-- also greedily maximizes the margins of the training examples. It allows much of the theory behind AdaBoost, including the general form of the scoring functions and their properties, to carry over to OnlineBoost.

\begin{theorem}
\label{thm:thm1}
OnlineBoost greedily minimizes the exponential loss of the margin $L(y,F(\mathbf{x})) = e^{-yF(\mathbf{x})}$ via stochastic gradient descent steps in the space of functions $F(\mathbf{x})$.
\end{theorem}
\begin{proof_sketch} 
See Appendix A.
\end{proof_sketch}

The scores of the form of Eq.(\ref{eq:scores}) assigned to the training examples can be expressed~\cite{NikolaouThesis} in terms of their corresponding margins as follows:
\begin{equation}
s({\bf{x}})=\begin{cases} \frac{1}{2}(1+margin({\bf{x}},y)) &\mbox{,~~~if } y = 1 \\
\frac{1}{2}(1-margin({\bf{x}},y)) & \mbox{,~~~if } y = -1. \end{cases}
\end{equation}
We see that as $margin({\bf{x}},y) \rightarrow 1$, the scores assume values $s({\bf{x}})\rightarrow 0$ for negative examples and $s({\bf{x}})\rightarrow 1$ for positive examples. In other words, maximizing the margins --something not only AdaBoost and OnlineBoost, but all boosting algorithms do, by virtue of minimizing monotonically decreasing loss functions of the margin-- forces the ensemble to learn to assign scores that tend to $0$ and $1$.

As for scores of the form of Eq.(\ref{eq:scores2}), they can be expressed~\cite{Edakunni2011} as a \emph{Product of Experts (PoE)}~\cite{Hinton2002}
%\begin{equation}
%S'_{T}({\bf{x}})=\frac{\prod_{\tau = 1}^{T}{s_{\tau}({\bf{x}})}}{\prod_{\tau = 1}^{T}{s_{\tau}({\bf{x}})} + \prod_{\tau = 1}^{T}{(1-s_{\tau}({\bf{x}}))}}.
%\end{equation}

%\hat{p}(y=1|{\bf{x}}) ~&=&~

\begin{eqnarray}
\label{eq:PoE_Ada}
s'({\bf{x}}) = \frac{\prod_{t=1}^{T} \hat{p}_t(y=1|{\bf{x}})}{\prod_{t=1}^{T} \hat{p}_{t}(y=1|{\bf{x}})+\prod_{t=1}^{T} \hat{p}_{t}(y=-1|{\bf{x}}) }, \nonumber 
\end{eqnarray}
\emph{with experts' probability estimates of the form}
\begin{eqnarray*}
\hat{p}_{t}(y=1|{\bf{x}}) = \begin{cases} \epsilon_{t}~~~~~, & \mbox{ if } h_{t}({\bf{x}}) = -1 \\ 1-\epsilon_{t}, & \mbox{ if } h_{t}({\bf{x}}) = 1, \end{cases}
\end{eqnarray*}
\begin{eqnarray*}
\hat{p}_{t}(y=-1|{\bf{x}}) = \begin{cases} 1-\epsilon_{t}, & \mbox{ if } h_{t}({\bf{x}}) = -1 \\ \epsilon_{t}~~~~~, & \mbox{ if } h_{t}({\bf{x}}) = 1, \end{cases}
\end{eqnarray*}
\emph{where $\epsilon_t = \frac{\lambda^{sw}_t}{\lambda^{sw}_t + \lambda^{sc}_t}$ is the weighted error of the $t$-th weak learner on the examples seen so far, and $h_{t}({\bf{x}}) \in\{ -1,1\}$ its prediction on example ${\bf{x}}$.}

In larger ensembles, the outputs tend to be more and more distorted. One reason for this is that the PoE assumes that the experts produce independent estimates and the more experts we add to the ensemble, the more likely we are to deviate from such an assumption. Another reason is that a single expert producing a score of $\hat{p}_t(y=1|{\bf{x}}) = 0$ or $\hat{p}_t(y=1|{\bf{x}}) = 1$ to a given example ${\bf{x}}$ suffices to dominate the ensemble's score $s'({\bf{x}})$ on that example. This was discussed in the case of batch AdaBoost in~\cite{Nikolaou2016} and holds for the OnlineBoost ensembles as well.

\begin{figure}\label{fig:uncalibrated_estimates}%[H]
\centering
\subfigure{\includegraphics[width=0.485\textwidth]{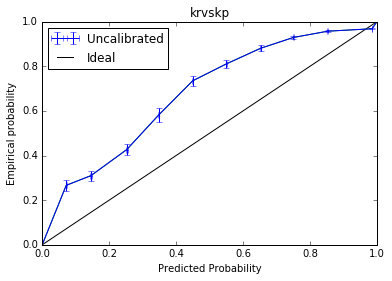}
  \label{}}
  %\vspace{-3.75cm}
%\bigskip
  %\vspace{-5cm}
\subfigure{\includegraphics[width=0.485\textwidth]{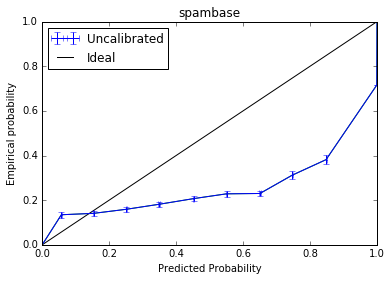}
  \label{}}
  %\vspace{-3.75cm}
%\bigskip
  %\vspace{-5cm}
\subfigure{\includegraphics[width=0.485\textwidth]{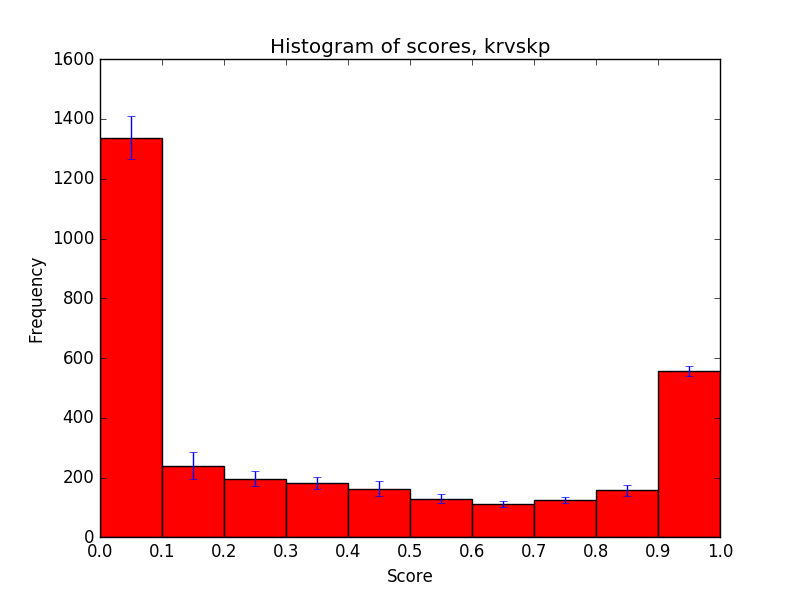}
  \label{}}
  %\vspace{-2.5cm}
%\bigskip
%\vspace{-1cm}
\subfigure{\includegraphics[width=0.485\textwidth]{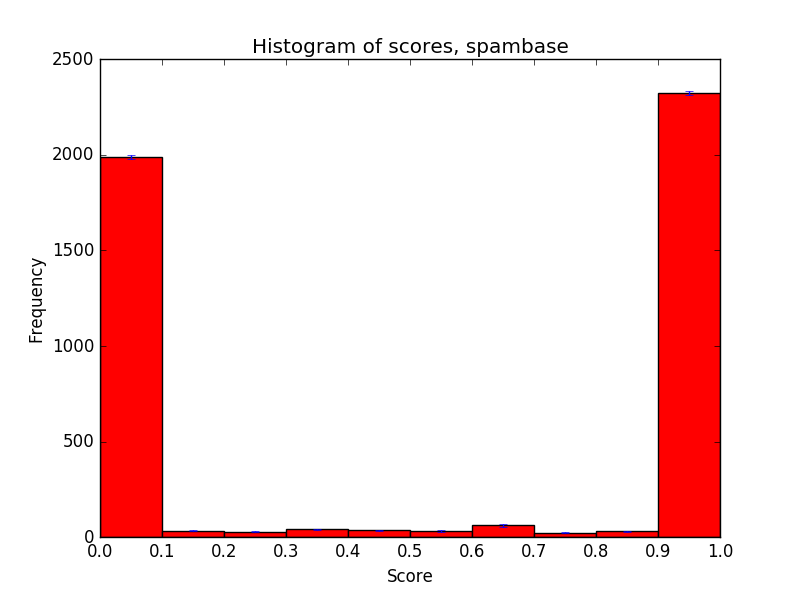}
  \label{}}
\caption{[TOP] Reliability diagrams for OnlineBoost ensembles on two sample datasets. The main diagonal corresponds to the ideal (perfectly calibrated) probability estimator. We can see that the probability estimates generated by OnlineBoost are far from ideal. [BOTTOM] Histogram of the probability estimates (scores) assigned by the boosting ensemble. We see the scores tend to values close to $0$ and $1$. The results shown here are averages and $95\%$ confidence intervals calculated across $10$ runs of training an ensemble of $T=10$ online Naive Bayes classifiers using minibatch sizes of $b=50$.}
%\vspace{-0.75em}
\end{figure}

In our experiments we use scores of the form of Eq.~(\ref{eq:scores}), motivated by previous work in batch boosting~\cite{NiculescuCaruana2005, CossockZhang2008, Nikolaou2016}. Indeed, as we see in Figure~\ref{fig:uncalibrated_estimates}, the scores tend to be skewed towards $0$ or $1$, and OnlineBoost ensembles tend to be very poorly calibrated. Note that poor probability estimation does not necessarily lead to poor classification. In fact, as we discussed, the very reason that makes boosting a successful classifier, namely its margin maximization property, is also responsible for its poor performance as a probability estimator, since it forces the ensemble to produce probability estimates skewed towards $0$ or $1$. %Note that given a sufficient amount of training data, calibrating the probability estimates cannot harm classification performance, as better probability estimates always result in better or at least the same predictions.
 
\section{Naive Calibration of Online Boosting}
In the previous section, we saw theoretical and empirical evidence that suggests that the probability estimates generated under OnlineBoost ensembles are distorted and need to be properly calibrated. To our knowledge, the calibration of online boosting ensembles has not been studied before in the literature. In this section we present a simple calibration policy, which directly draws from previous work on batch learning~\cite{NiculescuCaruana2005, Nikolaou2016}. In the next section we will further refine this approach.

\subsection{Online Platt Scaling}
\label{ssec:calibration_online}
As a calibration method we choose Platt-scaling (logistic calibration). This was done in part because OnlineBoost will tend to generate probability estimates that tend towards $0$ or $1$ as discussed above -i.e. amenable to sigmoid correction- and in part because the adaptation of the method in the online scenario is easy and efficient. Platt-scaling consists of finding a logistic sigmoid mapping of scores to probability estimates. The probability estimates are thus given by:
\begin{equation}
\hat{p}(y=1|{\bf{x}}) = \frac{1}{1+e^{w_{1}s({\bf{x}})+w_0}},
\end{equation}
where $s({\bf{x}})$ are the uncalibrated scores of the form of Eq.~(\ref{eq:scores}) and $w_0$ and $w_1$ are the parameters to be fitted.

We update the parameters of the sigmoid on one minibatch at a time (provided said minibatch is used for calibration --see next subsection), such that the log-loss of Eq.~(\ref{eq:logloss}) is minimized\footnote{In the paper we use log-loss to assess probabilistic predictions. Had we been using some other scoring rule (e.g. Brier score), it would be sensible to minimize the same loss (e.g. squared loss) to train the parameters of the sigmoid.}. To account for class imbalance, the Bayesian prior correction proposed by Platt~\cite{Platt1999} was applied. Rather than using $y_i=1$ for positive labels and $y_i=0$ for negative labels, in Eq.~(\ref{eq:logloss}), we use respectively
\begin{equation}
\label{eq:eq:y_pos_neg_logloss_cal}
y_i = \frac{N_{+}+1}{N_{+}+2}
   \quad\text{and}\quad 
y_i = \frac{1}{N_{-}+2}.
\end{equation}

On every minibatch (be it used for calibration or for training), we update $N_{+}$ and $N_{-}$, the current numbers of positive and negative examples respectively encountered in the dataset so far. If the data distribution is non-stationary, this also allows the predictions to adapt to prior probability shift.

Hence the calibrator always keeps track of $4$ quantities: $N_{+}$ and $N_{-}$ mentioned above and $w_0$ and $w_1$, the current sigmoid parameters to be updated on the next iteration. %The extra memory cost is constant as is desirable in an online setting. The fitting of $A$, $B$ is also cheap and can be done with any online optimizer.

\subsection{A Naive Calibration Policy}

A simple strategy is to use every $N_c$-th example to calibrate (update $w_0$ and $w_1$) and the remaining ones to train the ensemble\footnote{The first round is always used to train the ensemble.}. As the calibrator function has only two parameters, it is reasonable to expect that in the long term using most minibatches to train the ensemble would yield better results. We experiment with values $N_c \in \{ 2, 4, 6, 8, 10, 12, 14\} $, corresponding to fractions of $\{ 50\%, 25\%, 16.7\% ,12.5\%, 10\%, 8.3\%, 7.1\% \}$ of the data, respectively. Figure~2 shows reliability diagrams for uncalibrated OnlineBoost and naively-calibrated OnlineBoost with $N_c=2$ on two sample datasets. We see that even this most naive calibration policy considerably improves the probability estimation behaviour of online boosting. In the next section we will refine the naive calibration proposed here.

\begin{figure}\label{fig:some_calibration_better_than_none}%[H]
\centering
\subfigure{\includegraphics[width=0.485\textwidth]{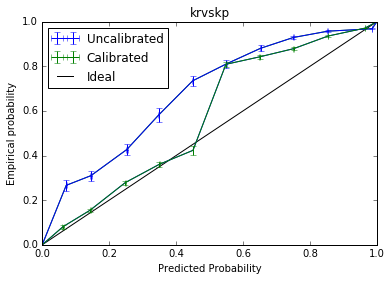}
  \label{}}
  %\vspace{-3.75cm}
%\bigskip
  %\vspace{-5cm}
\subfigure{\includegraphics[width=0.485\textwidth]{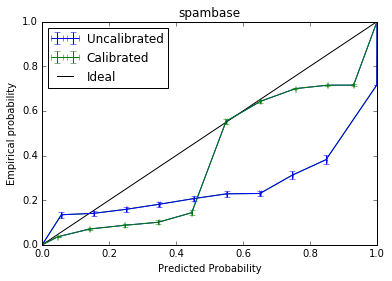}
  \label{}}
\caption{Reliability diagrams for uncalibrated OnlineBoost (green) and naively-calibrated OnlineBoost with $N_c=2$ (blue) on two sample datasets. The main diagonal corresponds to the ideal (perfectly calibrated) probability estimator. We can see that some calibration --even of the most naive form-- considerably improves the probability estimation behaviour of online boosting. The results shown here are averages and $95\%$ confidence intervals calculated across $10$ runs of training an ensemble of $T=10$ online Naive Bayes classifiers using minibatch sizes of $b=50$.}
\end{figure}

For completeness we should mention another obvious candidate naive policy of calibration: On each minibatch, we can construct two models, one by performing each of the two actions (train, calibrate). We then retain the model that leads to the greatest decrease in log-loss. This approach is computationally costlier, but as we only have two possible actions the increase in computational cost is constant and potentially affordable. However, when processing large amounts of data even a constant increase in computational cost can matter. Another flaw is that this approach is less amenable to extensions; in the next section we will discuss a bandit-based approach that can scale to an arbitrary number of other actions besides the two discussed here, or that can easily be adjusted to deal with non-stationary data, adversarial environments or incorporate contextual information.

Most importantly, however, the policy is memoryless and greedy. It does not encourage exploration and only exploits the action that reduced the log-loss the most on the last minibatch. Considering that unlike updating the ensemble parameters (which are geared towards reducing the classification error, leaving the final estimate poorly calibrated), the update of the calibrator parameters is explicitly performed with the objective of minimizing the log-loss, that action is almost guaranteed to be `calibrate'. Indeed, in our experiments with this technique, we saw that it reduced to always choosing to update the parameters of the calibrator. This caused the final probability estimates generated by this policy to be far worse than those produced by the other policies discussed here. We will therefore exclude it from further consideration.

\section{Bandit Algorithms for Calibrated Online Boosting}
\label{sec:bandits_for_cal}
Our results showed that employing a naive calibration policy is preferable to not calibrating the probability estimates at all. It is clear however, that the success of such a policy will depend on many factors: (1) the characteristics of the dataset, (2) the evaluation measure used (log-likelihood or Brier score for probability estimation, empirical risk for cost-sensitive classification, etc.), (3) the hyperparameters of the ensemble (weak learner, number of weak learners, scoring function), (4) the hyperparameters of the calibrator (choice of calibration function, optimization algorithm to train it). The interplay of these will determine the best value of $N_c$ for a fixed policy of the type discussed in the previous section.

Ideally, we would like to \emph{automate} the process of learning a good policy of alternating between the two actions. This is already an issue in batch learning, where determining the correct fraction of the training data that will be used for calibration is not obvious and will depend on all the factors mentioned in the previous paragraph. But in an online setting, it becomes even more important. The value of $N_c$ cannot be determined by cross-validation as predictions need to be made on the fly. Moreover, the optimal value of $N_c$ might \emph{change} during execution due to possible non-stationarity of the data (be it stochastic or adversarial).

To solve this problem we employed the different bandit optimization algorithms described in Section~\ref{sec:bandits}. The general methodology is described in Algorithm~\ref{alg:BanditCalibrate}. Each of the two actions (train, calibrate) is associated with a reward distribution. After each action is taken, the parameters of its corresponding reward distribution are updated accordingly.

\begin{algorithm}[htb]
   \caption{Deciding when to calibrate under a given $BanditPolicy$}
   \label{alg:BanditCalibrate}
\begin{algorithmic}
\STATE  {\bfseries For each round $n$ do:}
\STATE  {1} \ \ \ \ \ Receive unlabelled examples of $Minibatch_n$
\STATE  {2} \ \ \ \ \ Make predictions on examples of $Minibatch_n$
\STATE  {3} \ \ \ \ \ Receive labels of examples of $Minibatch_n$
\STATE  {4} \ \ \ \ \ Evaluate performance on $Minibatch_n$
\STATE  {5} \ \ \ \ \ {\bfseries If $n<2$ do}:
\STATE  {6} \ \ \ \ \ \ \ \ \ \ Update ensemble parameters on examples of $Minibatch_n$ (Sec.~\ref{ssec:boosting})
\STATE  {7} \ \ \ \ \ {\bfseries Else}: 
\STATE  {8} \ \ \ \ \ \ \ \ \ \ Use reward $X_n$ to update the parameters of $BanditPolicy$ 
\STATE  {9} \ \ \ \ \ \ \ \ \ \ Use $BanditPolicy$ to decide action $a_{n}$ (Sec.~\ref{sec:bandits})
\STATE  {10} \ \ \ \ \ \ \ \ \ {\bfseries If $a_{n} ==$} \emph{`TRAIN'} {\bfseries do}: 
\STATE  {11} \ \ \ \ \ \ \ \ \ \ \ \ \ \ Update ensemble parameters on examples of $Minibatch_n$ (Sec.~\ref{ssec:boosting})
\STATE  {12} \ \ \ \ \ \ \ \ \ {\bfseries Else}:
\STATE  {13} \ \ \ \ \ \ \ \ \ \ \ \ \ \ Update calibrator parameters on examples of $Minibatch_n$ (Sec.~\ref{ssec:calibration_online})
\STATE  {14} \ \ \ \ Compute reward $X_n$ of performing action $a_{n}$ (Sec.~\ref{sec:bandits_for_cal})
\end{algorithmic}
\end{algorithm}

The reward for each action is defined as the resulting \emph{relative decrease in log-loss after the action is taken}:
\begin{equation}
X_n(a_{n}) = -\frac{\mathcal{L}_n - \mathcal{L}_{n-1}}{\mathcal{L}_{n-1}} =  \frac{\mathcal{L}_{n-1} - \mathcal{L}_n}{\mathcal{L}_{n-1}} = 1 - \frac{\mathcal{L}_{n}}{\mathcal{L}_{n-1}} \in \mathbb{R},
\end{equation}
where $\mathcal{L}_{n-1}$ is the log-loss of round ${n-1}$, i.e. before performing action $a_{n}$ and $\mathcal{L}_{n}$ is the log-loss of round ${n}$, i.e. after performing action $a_{n}$.

For the cases of UCB1, UCB1-Improved and Gaussian Thompson Sampling, we also implemented versions employing discounted rewards to deal with the potentially non-stationary nature of online learning. More specifically, on each update, the cumulative rewards are multiplied by a discounting factor $\gamma<1$, i.e. $R_n = \gamma R_{n-1} +r_n$. The result is that the influence of past rewards decays geometrically.
 
We should note here that the non-stationarity can be due to the distribution of the data changing, but also due to the actions performed, which might lead to the reward distributions of the two actions changing. For example, after many rounds of only performing one action (e.g. `training'), we would intuitively expect that the reward distributions of the two actions have changed considerably, the rewards for training becoming smaller and smaller and the actual reward of calibrating having increased considerably since last sampled. Especially the model we have for an action not taken for many rounds (`calibrate', in this example) is expected to be poor. Discounting can protect us to some extent from such behaviours.

%[TODO: Could mention GPTS/Expected Improvement once finished and tested]

\section{Empirical Evaluation}
\subsection{Experimental Setup}
In our experiments we compared uncalibrated OnlineBoost to its naively calibrated version --i.e. a fixed policy of calibrating every $N_c$ rounds-- with $N_c \in \{ 2, 4, 6, 8, 10, 12, 14\}$, as well as to calibration under UCB1, UCB1-Improved and Gaussian Thompson Sampling policies and their discounted counterparts.

The uncalibrated probability estimates were of the form of Eq.~(\ref{eq:scores}). In the calibrated variants, logistic calibration was applied, by minimizing the loss of Eq.~(\ref{eq:logloss}), with incremental BFGS steps\footnote{Although BFGS is not typically a very popular choice for online learning, as the calibration step here always consists of updating the two parameters of a sigmoid, the computational and memory cost of BFGS is constant and low.}. %The initial values of the sigmoid parameters were set to [$A_0$, $B_0$] = [$0$, $log((N_{-} + 1.) / (N_{+} + 1.)$], where N_{+} and N_{-} is the number of positive and negative examples of the first minibatch.

We experimented with different choices of weak learners, both lossless (Gaussian Naive Bayes) and lossy (logistic regression, linear SVM, perceptron --all trained with stochastic gradient descent)\footnote{In the case of Gaussian Naive Bayes we used the resampling version of OnlineBoost given in Algorithm \ref{alg:OnlineBoost} --i.e. the original from \cite{oza2005}. For all other learners, we used the faster reweighting version described in Footnote $3$ as they supported it.}. We also examined the effect of different ensemble sizes ($T\in \{10,25,50\}$) and explored different degrees of regularization on the weak learner ($\ell_1$-regularized logistic regression with a regularization parameter $\lambda \in \{10^{-1}, 10^{-2}, 10^{-3},0 \}$). Unless otherwise specified the default parameters of \emph{scikit-learn}\footnote{http://scikit-learn.org/stable/} were used.

The hyperparameters of the bandit algorithms were fixed, as the purpose of using these algorithms is to circumvent hyperparameter tuning. The Gaussian prior for Thompson Sampling was set to $\mathcal{N}(0,1)$. Discounted reward versions used a discount factor $\gamma = 0.95$.

%\footnote{In the case of Gaussian Naive Bayes we used the resampling version of OnlineBoost given in Algorithm \ref{alg:OnlineBoost} (i.e. the original from \cite{oza2005}). For all other learners we used the faster reweighting approach described in Footnote $3$ as they supported it.}, both lossless (Gaussian Naive Bayes) and lossy (logistic regression, linear SVM, perceptron), we also examined the effect of different ensemble sizes ($T \in \{10, 25, 50\}$) and explored different degrees of regularization on the weak learner ($l1$-regularized logistic regression with regularization parameter $\lambda \in \{10^{-1}, 10^{-2}, 10^{-3},0 \}$).

The experiments were carried out on $10$ real-world datasets, the characteristics of which are given in Appendix $B$. The examples in \emph{krvskp}, \emph{landsat}, \emph{splice}, \emph{waveform}, \emph{spambase}, \emph{mushroom}, \emph{musk2} are considered i.i.d., so they are used for simulating situations where online learning is employed to generate good predictions faster than a batch learning algorithm. For these, the minibatch size was set to $b=50$. The datasets \emph{weather}, \emph{electricity} and \emph{forest} were considered non-stationary. As these three datasets are also considerably larger than the other $7$, the minibatch size was set to $b=100$ for faster processing.

%For all experiments we used online Gaussian Naive Bayes as the base learner. It was chosen as it is a lossless on-line learner with a simple online implementation. The ensemble size was set to $T=25$ for all datasets.

\subsection{Experimental Results}
We present the negative log-likelihood across the entire dataset as an overall measure of performance of each variant. We present the best and worst result attained on average by fixed policies (in the sense of final log-loss attained) and specify in each case the corresponding $N_c$ that produced it. Only some characteristic results are presented here. The remaining ones are given in Appendix $C$.

We also provide some characteristic learning curves on the negative log-likelihood (average negative log-likelihood across all past predictions versus number of minibatches seen). This allows us to observe how fast each algorithm can generate good probability estimates. In each case, we report average values and $95\%$ confidence intervals across $10$ runs\footnote{Datasets that are i.i.d. are shuffled on each run, thus changing the order in which examples are presented to the learner. On non-stationary datasets, we respect the order in which the examples arrive to preserve their non-stationary nature.}. 

\subsubsection{Experiments on stationary datasets.} 
\begin{figure}\label{fig:NB10_stationary}%[H]
\centering
\subfigure{\includegraphics[width=0.485\textwidth]{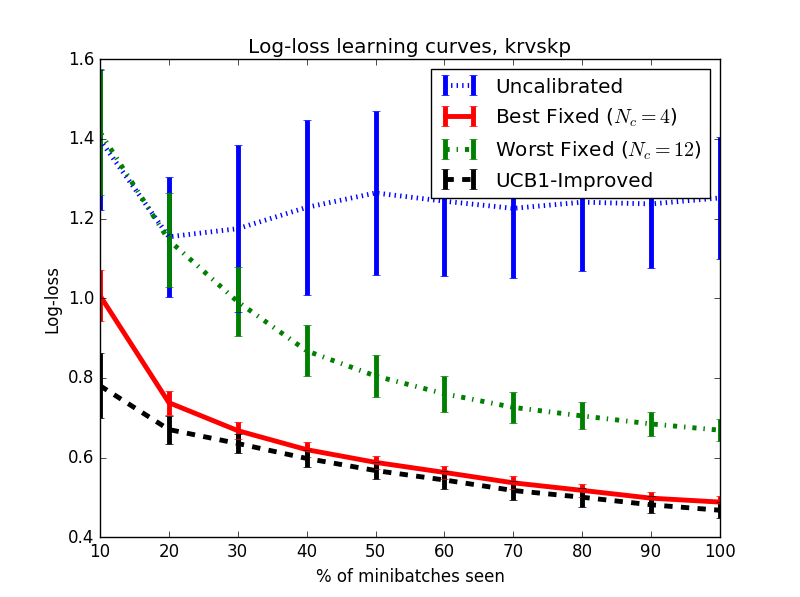}
  \label{}}
  %\vspace{-3.75cm}
%\bigskip
  %\vspace{-5cm}
%\subfigure{\includegraphics[width=0.485\textwidth]{./Figures/NB10_krvskp_discGTS}
 % \label{}}
  %\vspace{-3.75cm}
%\bigskip
  %\vspace{-5cm}
\subfigure{\includegraphics[width=0.485\textwidth]{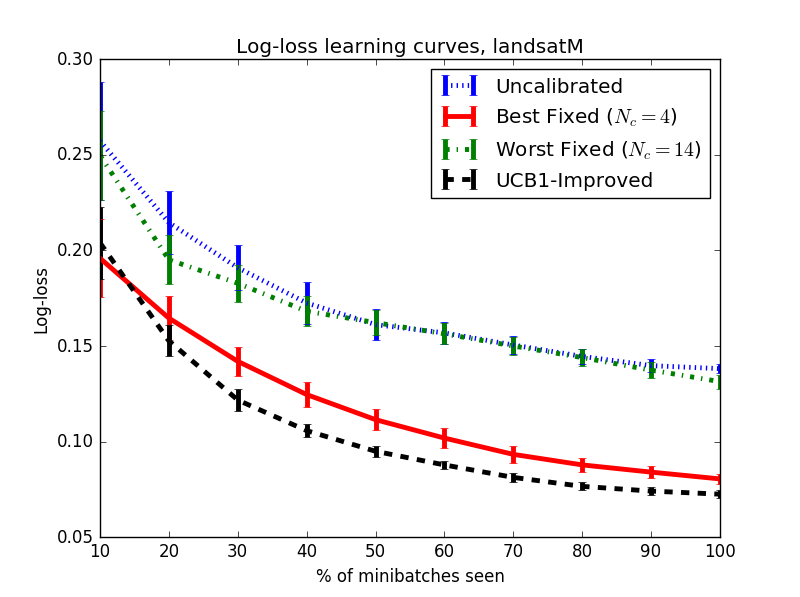}
  \label{}}
  %\vspace{-2.5cm}
%\bigskip
%\vspace{-1cm}
%\subfigure{\includegraphics[width=0.485\textwidth]{./Figures/NB10_landsatM_discGTS}
%  \label{}}
%  \bigskip
  %\vspace{-5cm}
\subfigure{\includegraphics[width=0.485\textwidth]{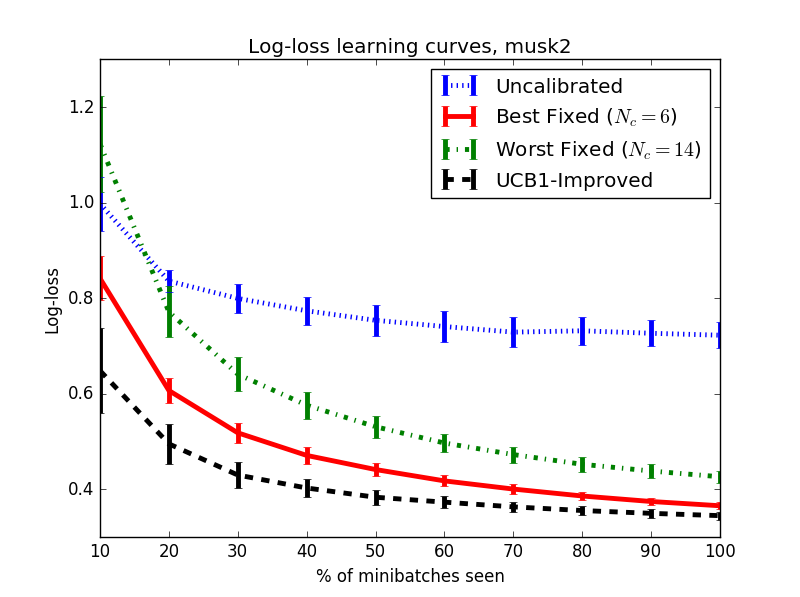}
  \label{}}
  %\vspace{-2.5cm}
%\bigskip
%\vspace{-1cm}
%\subfigure{\includegraphics[width=0.485\textwidth]{./Figures/NB10_musk2_discGTS}
%  \label{}}
%  \bigskip
  %\vspace{-5cm}
\subfigure{\includegraphics[width=0.485\textwidth]{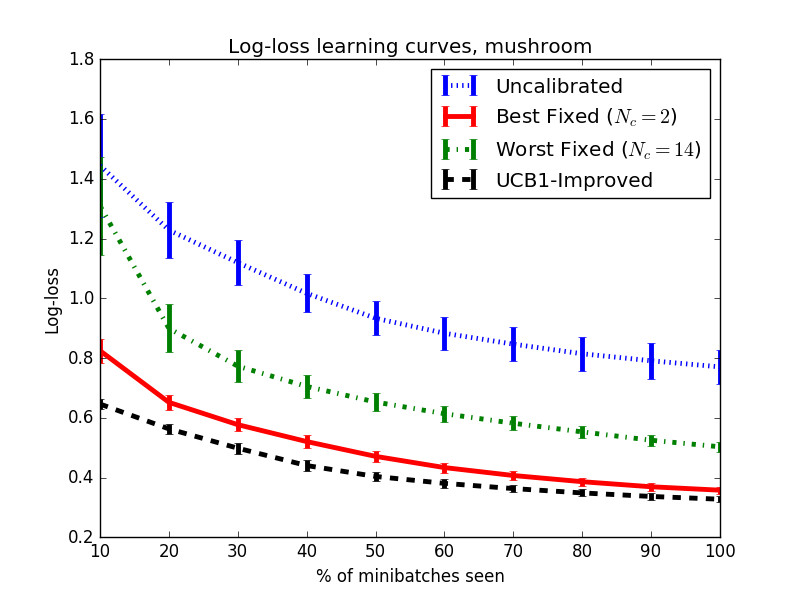}
  \label{}}
  %\vspace{-2.5cm}
%\bigskip
%\vspace{-1cm}
%\subfigure{\includegraphics[width=0.485\textwidth]{./Figures/NB10_mushroom_discGTS}
%  \label{}}
  %\vspace{-1cm}
\caption{Log-loss versus number of minibatches seen. Lower values correspond to better probability estimation. The best and worst fixed calibration policies are compared against uncalibrated OnlineBoost and to calibration under the non-discounted UCB1-Improved policy. UCB1, discounted Gaussian Thompson Sampling performed similarly. Results for Naive Bayes, $T=10$ on stationary datasets. Only datasets for which the best fixed policy is competitive with bandit policies are shown. Note how fast the log-loss reduces under the bandit policies.}
%\vspace{-0.75em}
\end{figure}

\begin{figure}\label{fig:LR10_stationary}%[H]
\centering
\subfigure{\includegraphics[width=0.485\textwidth]{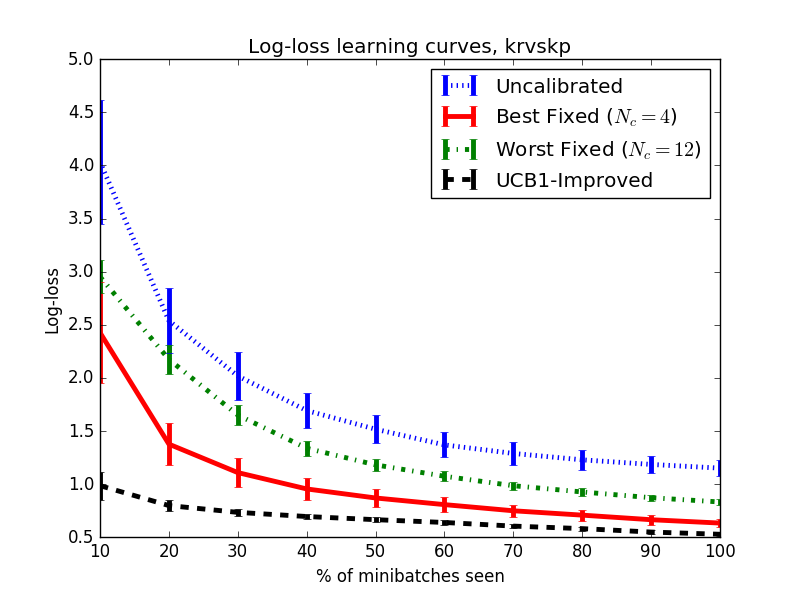}
  \label{}}
  %\vspace{-3.75cm}
%\bigskip
  %\vspace{-5cm}
%\subfigure{\includegraphics[width=0.485\textwidth]{./Figures/LR10_krvskp_DiscGTS}
  %\label{}}
  %\vspace{-3.75cm}
%\bigskip
  %\vspace{-5cm}
\subfigure{\includegraphics[width=0.485\textwidth]{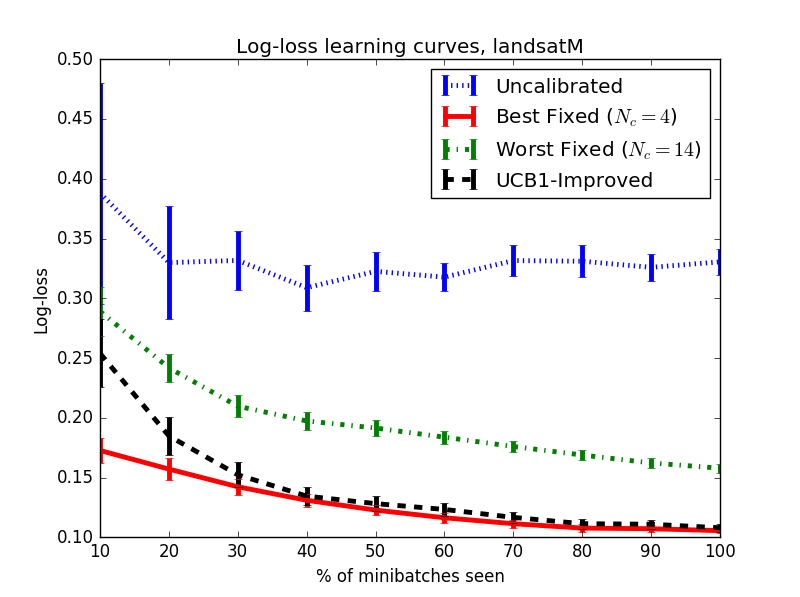}
  \label{}}
  %\vspace{-2.5cm}
%\bigskip
%\vspace{-1cm}
%\subfigure{\includegraphics[width=0.485\textwidth]{./Figures/LR10_landsatM_DiscGTS}
  %\label{}}
  %\bigskip
  %\vspace{-5cm}
\subfigure{\includegraphics[width=0.485\textwidth]{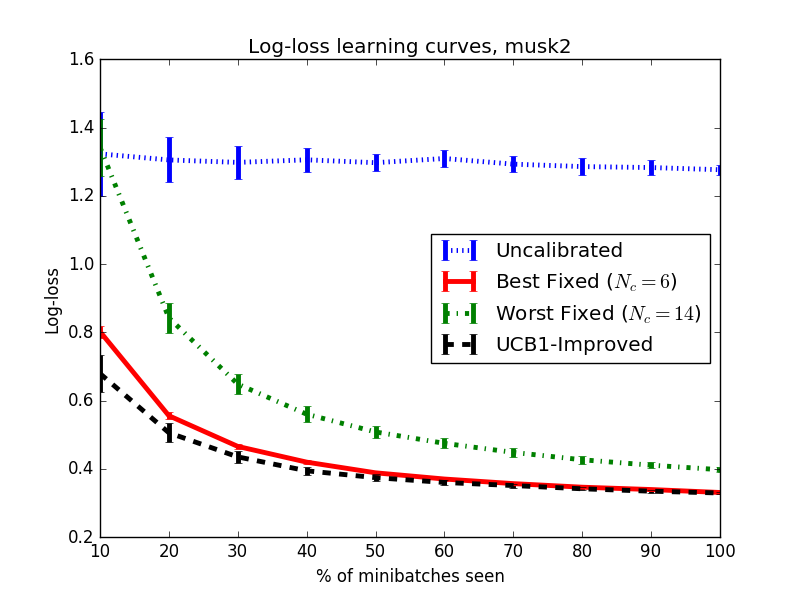}
  \label{}}
  %\vspace{-2.5cm}
%\bigskip
%\vspace{-1cm}
%\subfigure{\includegraphics[width=0.485\textwidth]{./Figures/LR10_musk2_DiscGTS}
%  \label{}}
  %\bigskip
  %\vspace{-5cm}
\subfigure{\includegraphics[width=0.485\textwidth]{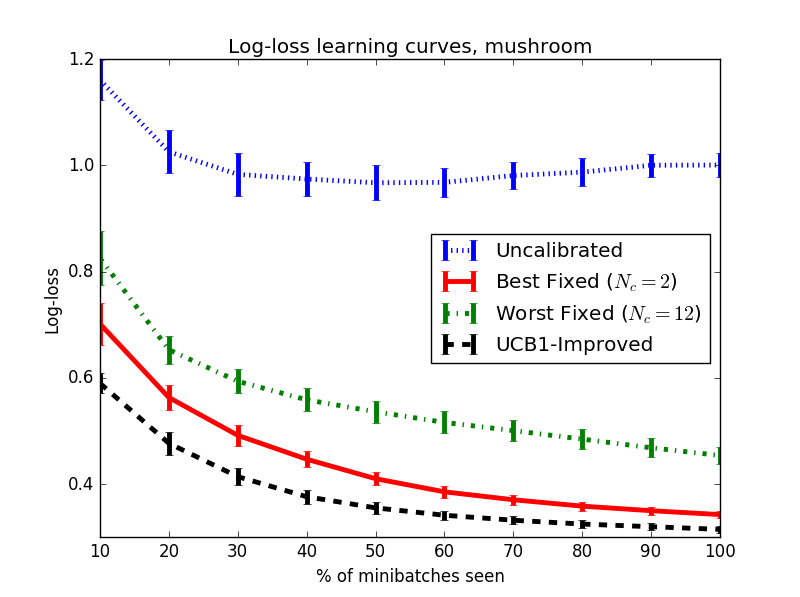}
  \label{}}
  %\vspace{-2.5cm}
%\bigskip
%\vspace{-1cm}
%\subfigure{\includegraphics[width=0.485\textwidth]{./Figures/LR10_mushroom_DiscGTS}
%  \label{}}
  %\vspace{-1cm}
\caption{Log-loss versus number of minibatches seen. Lower values correspond to better probability estimation. The best and worst fixed calibration policies are compared against uncalibrated OnlineBoost and calibration under the non-discounted UCB1-Improved policy. UCB1, discounted Gaussian Thompson Sampling performed similarly. Results for logistic regression, $T=10$ on stationary datasets. Only datasets for which the best fixed policy is competitive with bandit policies are shown. Note how fast the log-loss reduces under the bandit policies.}
%\vspace{-0.75em}
\end{figure}

We shall first present the results under various choices of weak learners on stationary datasets. The results for varying ensemble size and degree of regularization can be found in Appendix $C$, as they are qualitatively similar. Tables \ref{tab:NB10_stationary_AAA}--\ref{tab:PERC10_stationary_AAA} show the log-loss across the entire dataset. Figures \ref{fig:NB10_stationary} \& \ref{fig:LR10_stationary} show the evolution of average log-loss during training for some selected combinations of dataset, base learner and policy. 

\begin{table}[H]
%\scriptsize
\center
\caption{Log-loss across the entire dataset. Lowest average value shown in bold. Results for Naive Bayes, $T=10$ on stationary datasets.}\label{tab:NB10_stationary_AAA}
%\vspace{-0.2cm}
    \begin{tabular}{|c|c|c|c|c|c|c|c|c|c|}
    \hline
 	\multirow{3}{*}{Dataset} &   & Best & Worst &  &  UCB1 &   & Disc. & Disc. & Disc. \\
 	 & Uncalibrated & Fixed & Fixed & UCB1 & Improved & GTS & UCB1 & UCB1 & GTS\\ 
     &  &  &  &  &  &  &  & Improved & \\ \hline\hline    
        \multirow{3}{*}{\emph{landsat}} & $0.138$ & $0.081$ & $0.131$ & $0.072$ & $\mathbf{0.071}$ & $0.074$ & $0.129$ &  $0.123$ &  $0.073$ \\
       & $\pm$ & $\pm$ & $\pm$ & $\pm$ & $\mathbf{\pm}$ & $\pm$ & $\pm$ & $\pm$ & $\pm$ \\
       & $0.002$ & $0.002$ & $0.004$ & $0.001$ & $\mathbf{0.001}$ & $0.002$ & $0.010$ & $0.011$ & $0.002$ \\ \hline              
    \multirow{3}{*}{\emph{splice}} & $0.444$ & $0.280$ & $0.410$ & $0.234$ & $\mathbf{0.229}$ & $0.282$ & $0.353$ & $0.376$ & $0.232$ \\ 
     & $\pm$ & $\pm$ & $\pm$ & $\pm$ & $\mathbf{\pm}$ & $\pm$ & $\pm$ & $\pm$ & $\pm$ \\
     & $0.034$ & $0.014$ & $0.014$ & $0.007$ & $\mathbf{0.009}$ & $0.022$ & $0.024$ & $0.037$ & $0.008$ \\ \hline              
    \multirow{3}{*}{\emph{musk2}} & $0.723$ & $0.343$ & $0.427$ & $\mathbf{0.320}$ & $0.332$ & $0.397$ & $0.711$ & $0.441$ & $0.345$ \\ 
     & $\pm$ & $\pm$ & $\pm$ & $\pm$ & $\mathbf{\pm}$ & $\pm$ & $\pm$ & $\pm$ & $\pm$ \\
     & $0.027$ & $0.008$ & $0.013$ & $\mathbf{0.006}$ & $0.003$ & $0.008$ & $0.025$ & $0.049$ & $0.008$ \\ \hline              
    \multirow{3}{*}{\emph{krvskp}} & $1.252$ & $0.488$ & $0.720$ & $0.474$ & $0.472$ & $0.486$ & $0.627$ & $0.767$ & $\mathbf{0.468}$ \\
       & $\pm$ & $\pm$ & $\pm$ & $\pm$ & $\pm$ & $\pm$ & $\pm$ & $\pm$ & $\mathbf{\pm}$ \\
       & $0.154$ & $0.014$ & $0.075$ & $0.019$ & $0.012$ & $0.016$ & $0.114$ & $0.114$ & $\mathbf{0.020}$ \\ \hline    
    \multirow{3}{*}{\emph{waveform}} & $0.824$ & $0.361$ & $0.466$ & $0.335$ & $\mathbf{0.334}$ & $0.342$ & $0.458$ & $0.492$ & $0.344$\\
       & $\pm$ & $\pm$ & $\pm$ & $\pm$ & $\pm$ & $\pm$ & $\pm$ & $\pm$ & $\pm$ \\
       & $0.012$ & $0.005$ & $0.009$ & $0.002$ & $\mathbf{0.004}$ & $0.007$ & $0.015$ & $0.024$ & $0.006$ \\ \hline              
    \multirow{3}{*}{\emph{spambase}} & $2.37$ & $0.532$ & $0.735$ & $0.493$ & $0.483$ & $\mathbf{0.481}$ & $0.536$ & $0.540$ & $0.489$\\
       & $\pm$ & $\pm$ & $\pm$ & $\pm$ & $\pm$ & $\mathbf{\pm}$ & $\pm$ & $\pm$ & $\pm$ \\
       & $0.026$ & $0.006$ & $0.017$ & $0.006$ & $0.005$ & $\mathbf{0.008}$ & $0.028$ & $0.037$ & $0.005$ \\ \hline                
    \multirow{3}{*}{\emph{mushroom}} & $0.770$ & $0.358$ & $0.503$ & $0.375$ & $0.335$ & $\mathbf{0.318}$ & $0.570$ & $0.570$ & $0.327$ \\ 
           & $\pm$ & $\pm$ & $\pm$ & $\pm$ & $\pm$ & $\pm$ & $\mathbf{\pm}$ & $\pm$ & $\pm$ \\
       & $0.057$ & $0.011$ & $0.017$ & $0.018$ & $0.020$ & $\mathbf{0.006}$ & $0.036$ & $0.072$ & $0.010$ \\ \hline  
    \end{tabular}
   % \vspace{-3cm}
\end{table}
%\vspace{-1.25cm}

\begin{table}[H]
\center
\caption{Log-loss across the entire dataset. Lowest average value shown in bold. Results for logistic regression, $T=10$ on stationary datasets. }\label{tab:LR10_stationary_AAA}
%\vspace{-0.2cm}
    \begin{tabular}{|c|c|c|c|c|c|c|c|c|c|}
    \hline
 	\multirow{3}{*}{Dataset} &   & Best & Worst &  &  UCB1 &   & Disc. & Disc. & Disc. \\
 	 & Uncalibrated & Fixed & Fixed & UCB1 & Improved & GTS & UCB1 & UCB1 & GTS\\ 
     &  &  &  &  &  &  &  & Improved & \\ \hline\hline    
        \multirow{3}{*}{\emph{landsat}} & $0.330$ & $\mathbf{0.106}$ & $0.158$ & $0.107$ & $0.108$ & $0.108$ & $0.173$ &  $0.203$ &  $0.127$ \\
       & $\pm$ & $\mathbf{\pm}$ & $\pm$ & $\pm$ & $\pm$ & $\pm$ & $\pm$ & $\pm$ & $\pm$ \\
       & $0.011$ & $\mathbf{0.002}$ & $0.004$ & $0.003$ & $0.004$ & $0.004$ & $0.019$ & $0.018$ & $0.013$ \\ \hline     
    \multirow{3}{*}{\emph{splice}} & $1.779$ & $0.626$ & $0.895$ & $\mathbf{0.506}$ & $0.554$ & $0.636$ & $0.711$ & $1.155$ & $0.584$ \\ 
     & $\pm$ & $\pm$ & $\pm$ & $\mathbf{\pm}$ & $\pm$ & $\pm$ & $\pm$ & $\pm$ & $\pm$ \\
     & $0.059$ & $0.023$ & $0.034$ & $\mathbf{0.029}$ & $0.020$ & $0.068$ & $0.117$ & $0.169$ & $0.034$ \\ \hline    
    \multirow{3}{*}{\emph{musk2}} & $1.276$ & $0.331$ & $0.398$ & $0.321$ & $0.330$ & $\mathbf{0.318}$ & $0.536$ & $0.514$ & $0.322$ \\ 
    & $\pm$ & $\pm$ & $\pm$ & $\pm$ & $\pm$ & $\mathbf{\pm}$ & $\pm$ & $\pm$ & $\pm$ \\
     & $0.015$ & $0.003$ & $0.008$ & $0.004$ & $0.004$ & $\mathbf{0.003}$ & $0.049$ & $0.038$ & $0.006$ \\ \hline       
    \multirow{3}{*}{\emph{krvskp}} & $1.151$ & $0.633$ & $0.881$ & $\mathbf{0.519}$ & $0.528$ & $0.645$ & $0.825$ & $0.733$ & $0.555$ \\
       & $\pm$ & $\pm$ & $\pm$ & $\mathbf{\pm}$ & $\pm$ & $\pm$ & $\pm$ & $\pm$ & $\pm$ \\
       & $0.073$ & $0.041$ & $0.042$ & $\mathbf{0.037}$ & $0.017$ & $0.048$ & $0.074$ & $0.041$ & $0.033$ \\ \hline 
    \multirow{3}{*}{\emph{waveform}} & $1.700$ & $0.444$ & $0.596$ & $\mathbf{0.427}$ & $0.429$ & $0.428$ & $0.503$ & $0.519$ & $0.464$\\
       & $\pm$ & $\pm$ & $\pm$ & $\pm$ & $\pm$ & $\pm$ & $\pm$ & $\pm$ & $\pm$ \\
       & $0.037$ & $0.004$ & $0.019$ & $\mathbf{0.003}$ & $0.007$ & $0.006$ & $0.026$ & $0.030$ & $0.016$ \\ \hline  
    \multirow{3}{*}{\emph{spambase}} & $1.182$ & $0.409$ & $0.572$ & $0.390$ & $0.395$ & $0.407$ & $0.540$ & $0.477$ & $\mathbf{0.389}$\\
       & $\pm$ & $\pm$ & $\pm$ & $\pm$ & $\pm$ & $\pm$ & $\pm$ & $\pm$ & $\mathbf{\pm}$ \\
       & $0.018$ & $0.006$ & $0.010$ & $0.005$ & $0.006$ & $0.007$ & $0.029$ & $0.024$ & $\mathbf{0.006}$ \\ \hline      
    \multirow{3}{*}{\emph{mushroom}} & $1.001$ & $0.343$ & $0.454$ & $0.322$ & $0.315$ & $\mathbf{0.312}$ & $0.531$ & $0.592$ & $0.330$ \\ 
           & $\pm$ & $\pm$ & $\pm$ & $\pm$ & $\pm$ & $\mathbf{\pm}$ & $\pm$ & $\pm$ & $\pm$ \\
       & $0.023$ & $0.007$ & $0.017$ & $0.007$ & $0.006$ & $\mathbf{0.007}$ & $0.026$ & $0.026$ & $0.006$ \\ \hline    
    \end{tabular}
   % \vspace{-3cm}
\end{table}
%\vspace{-1.25cm}

\begin{table}[H]
%\scriptsize
\center
\caption{Log-loss across the entire dataset. Lowest average value shown in bold. Results for linear SVM, $T=10$ on stationary datasets. }\label{tab:LINSVM10_stationary_AAA}
%\vspace{-0.2cm}
    \begin{tabular}{|c|c|c|c|c|c|c|c|c|c|}
    \hline
 	\multirow{3}{*}{Dataset} &   & Best & Worst &  &  UCB1 &   & Disc. & Disc. & Disc. \\
 	 & Uncalibrated & Fixed & Fixed & UCB1 & Improved & GTS & UCB1 & UCB1 & GTS\\ 
     &  &  &  &  &  &  &  & Improved & \\ \hline\hline    
        \multirow{3}{*}{\emph{landsat}} & $0.245$ & $ 0.130$ & $0.184$ & $0.118$ & $0.121$ & $0.127$ & $0.250$ & $0.243$ & $\mathbf{0.111}$ \\ 
           & $\pm$ & $\pm$ & $\pm$ & $\pm$ & $\pm$ & $\pm$ & $\pm$ & $\pm$ & $\mathbf{\pm}$ \\
       & $0.019$ & $0.009$ & $0.010$ & $0.008$ & $0.008$ & $0.010$ & $0.025$ & $0.034$ & $\mathbf{0.004}$ \\ \hline 
    \multirow{3}{*}{\emph{splice}} & $1.391$ & $0.663$ & $0.922$ & $0.622$ & $\mathbf{0.586}$ & $0.609$ & $0.816$ & $0.666$ & $0.640$ \\ 
           & $\pm$ & $\pm$ & $\pm$ & $\pm$ & $\mathbf{\pm}$ & $\pm$ & $\pm$ & $\pm$ & $\pm$ \\
       & $0.064$ & $0.018$ & $0.040$ & $0.031$ & $\mathbf{0.022}$ & $0.020$ & $0.105$ & $0.088$ & $0.038$ \\ \hline 
    \multirow{3}{*}{\emph{musk2}} & $0.953$ & $\mathbf{0.385}$ & $0.432$ & $0.400$ & $0.390$ & $0.406$ & $0.639$ & $0.622$ & $0.386$ \\ 
          & $\pm$ & $\mathbf{\pm}$ & $\pm$ & $\pm$ & $\pm$ & $\pm$ & $\pm$ & $\pm$ & $\pm$ \\
       & $0.020$ & $\mathbf{0.008}$ & $0.007$ & $0.006$ & $0.007$ & $0.008$ & $0.061$ & $0.066$ & $0.005$ \\ \hline          
    \multirow{3}{*}{\emph{krvskp}} & $1.081$ & $0.671$ & $0.897$ & $\mathbf{0.608}$ & $0.648$ & $0.677$ & $0.706$ & $0.811$ & $0.667$ \\ 
           & $\pm$ & $\pm$ & $\pm$ & $\mathbf{\pm}$ & $\pm$ & $\pm$ & $\pm$ & $\pm$ & $\pm$ \\
       & $0.050$ & $0.022$ & $0.027$ & $\mathbf{0.021}$ & $0.027$ & $0.029$ & $0.047$ & $0.041$ & $0.030$ \\ \hline 
    \multirow{3}{*}{\emph{waveform}} & $1.156$ & $0.487$ & $0.615$ & $0.453$ & $\mathbf{0.451}$ & $\mathbf{0.451}$ & $0.611$ & $0.603$ & $0.464$ \\ 
          & $\pm$ & $\pm$ & $\pm$ & $\pm$ & $\mathbf{\pm}$ & $\mathbf{\pm}$ & $\pm$ & $\pm$ & $\pm$ \\
       & $0.045$ & $0.013$ & $0.015$ & $0.005$ & $\mathbf{0.004}$ & $\mathbf{0.005}$ & $0.021$ & $0.076$ & $0.005$ \\ \hline 
    \multirow{3}{*}{\emph{spambase}} & $0.893$ & $0.440$ & $0.551$ & $0.411$ & $\mathbf{0.410}$ & $0.418$ & $0.560$ & $0.612$ & $0.414$ \\ 
          & $\pm$ & $\pm$ & $\pm$ & $\pm$ & $\pm$ & $\pm$ & $\pm$ & $\pm$ & $\pm$ \\
       & $0.019$ & $0.008$ & $0.010$ & $0.005$ & $\mathbf{0.007}$ & $0.009$ & $0.028$ & $0.053$ & $0.0037$ \\ \hline  
    \multirow{3}{*}{\emph{mushroom}} & $0.673$ & $0.374$ & $0.489$ & $0.359$ & $0.363$ & $\mathbf{0.352}$ & $0.506$ & $0.573$ & $0.357$ \\ 
           & $\pm$ & $\pm$ & $\pm$ & $\pm$ & $\pm$ & $\mathbf{\pm}$ & $\pm$ & $\pm$ & $\pm$ \\
       & $0.015$ & $0.006$ & $0.015$ & $0.006$ & $0.006$ & $\mathbf{0.005}$ & $0.036$ & $0.039$ & $0.007$ \\ \hline     
    \end{tabular}
   % \vspace{-3cm}
\end{table}
%\vspace{-1.25cm}

\begin{table}[H]
%\scriptsize
\center
\caption{Log-loss across the entire dataset. Lowest average value shown in bold. Results for perceptron, $T=10$ on stationary datasets. }\label{tab:PERC10_stationary_AAA}
%\vspace{-0.2cm}
    \begin{tabular}{|c|c|c|c|c|c|c|c|c|c|}
    \hline
 	\multirow{3}{*}{Dataset} &   & Best & Worst &  &  UCB1 &   & Disc. & Disc. & Disc. \\
 	 & Uncalibrated & Fixed & Fixed & UCB1 & Improved & GTS & UCB1 & UCB1 & GTS\\ 
     &  &  &  &  &  &  &  & Improved & \\ \hline\hline    
        \multirow{3}{*}{\emph{landsat}} & $0.258$ & $ 0.120$ & $0.190$ & $0.125$ & $\mathbf{0.103}$ & $0.115$ & $0.164$ & $0.198$ & $0.120$ \\ 
           & $\pm$ & $\pm$ & $\pm$ & $\pm$ & $\pm$ & $\pm$ & $\pm$ & $\pm$ & $\pm$ \\
       & $0.012$ & $0.005$ & $0.009$ & $0.013$ & $\mathbf{0.004}$ & $0.005$ & $0.014$ & $0.028$ & $0.008$ \\ \hline      
    \multirow{3}{*}{\emph{splice}} & $1.453$ & $0.683$ & $0.879$ & $\mathbf{0.566}$ & $0.624$ & $0.642$ & $0.759$ & $0.651$ & $0.607$ \\ 
           & $\pm$ & $\pm$ & $\pm$ & $\mathbf{\pm}$ & $\pm$ & $\pm$ & $\pm$ & $\pm$ & $\pm$ \\
       & $0.037$ & $0.033$ & $0.050$ & $\mathbf{0.021}$ & $0.028$ & $0.025$ & $0.105$ & $0.022$ & $0.027$ \\ \hline        
    \multirow{3}{*}{\emph{musk2}} & $0.945$ & $0.392$ & $0.443$ & $0.390$ & $\mathbf{0.385}$ & $0.419$ & $0.591$ & $0.648$ & $0.408$ \\ 
          & $\pm$ & $\pm$ & $\pm$ & $\pm$ & $\mathbf{\pm}$ & $\pm$ & $\pm$ & $\pm$ & $\pm$ \\
       & $0.030$ & $0.006$ & $0.012$ & $0.006$ & $\mathbf{0.006}$ & $0.014$ & $0.044$ & $0.067$ & $0.013$ \\ \hline 
    \multirow{3}{*}{\emph{krvskp}} & $1.021$ & $0.745$ & $0.909$ & $0.632$ & $\mathbf{0.609}$ & $0.777$ & $0.812$ & $0.811$ & $0.611$ \\ 
           & $\pm$ & $\pm$ & $\pm$ & $\pm$ & $\mathbf{\pm}$ & $\pm$ & $\pm$ & $\pm$ & $\pm$ \\
       & $0.042$ & $0.031$ & $0.043$ & $0.037$ & $\mathbf{0.022}$ & $0.079$ & $0.069$ & $0.062$ & $0.017$ \\ \hline 
    \multirow{3}{*}{\emph{waveform}} & $1.106$ & $0.498$ & $0.589$ & $0.468$ & $0.492$ & $0.486$ & $0.573$ & $0.686$ & $\mathbf{0.456}$ \\ 
          & $\pm$ & $\pm$ & $\pm$ & $\pm$ & $\pm$ & $\pm$ & $\pm$ & $\pm$ & $\pm$ \\
       & $0.027$ & $0.010$ & $0.021$ & $0.010$ & $0.034$ & $0.026$ & $0.029$ & $0.066$ & $\mathbf{0.008}$ \\ \hline 
    \multirow{3}{*}{\emph{spambase}} & $0.885$ & $0.430$ & $0.575$ & $\mathbf{0.400}$ & $0.415$ & $0.415$ & $0.521$ & $0.620$ & $0.418$ \\ 
          & $\pm$ & $\pm$ & $\pm$ & $\mathbf{\pm}$ & $\pm$ & $\pm$ & $\pm$ & $\pm$ & $\pm$ \\
       & $0.012$ & $0.006$ & $0.010$ & $\mathbf{0.004}$ & $0.004$ & $0.007$ & $0.027$ & $0.043$ & $0.008$ \\ \hline     
    \multirow{3}{*}{\emph{mushroom}} & $0.670$ & $0.383$ & $0.489$ & $0.359$ & $\mathbf{0.350}$ & $0.367$ & $0.502$ & $0.599$ & $0.355$ \\ 
           & $\pm$ & $\pm$ & $\pm$ & $\pm$ & $\mathbf{\pm}$ & $\pm$ & $\pm$ & $\pm$ & $\pm$ \\
       & $0.012$ & $0.006$ & $0.011$ & $0.003$ & $\mathbf{0.006}$ & $0.004$ & $0.032$ & $0.033$ & $0.006$ \\ \hline     
    \end{tabular}
   % \vspace{-3cm}
\end{table}
%\vspace{-1.25cm}

%Naive Bayes, being lossless, exhibits lower variance in its results than the other learners.

In Tables \ref{tab:NB10_stationary_AAA}--\ref{tab:PERC10_stationary_AAA} we show on bold the best policy on average for each dataset. Note that in many situations the confidence intervals overlap, in which situation no clear winning policy can be determined.   

We see that regardless of the weak learner used, it is almost always the case that applying some calibration (even under the worst fixed policy) produces significantly better probability estimates than applying no calibration. This is in vein with results batch boosting~\cite{NiculescuCaruana2005, Nikolaou2016} and agrees with both our theoretical intuitions and the empirical analysis of Section~\ref{sec:prob_est}. 

Moreover, it is almost always the case that certain bandit policies --more specifically the non-discounted versions of UCB1 and UCB1-Improved closely followed by Gaussian Thompson Sampling, especially in its discounted version-- significantly outperform the best fixed calibration policy. Even when they don't, the best fixed calibration policy does not significantly outperform them. We can conclude that these policies are producing at least as good probability estimates as the best fixed policy in each case.

To get a clearer picture of this, and see how fast the log-loss reduces under each policy, in Figures \ref{fig:NB10_stationary} \& \ref{fig:LR10_stationary} we provide some characteristic learning curves. As the bandit policies produce comparably good results, to prevent cluttering, we only include one bandit policy per figure (UCB1-Improved or discounted Gaussian Thompson Sampling), compared against the best and worst fixed policy and the uncalibrated online boosting ensemble. We specifically chose to visualize the results on the datasets for which the best fixed policy is competitive --for some choice of weak learner-- with bandit policies when its predictions are evaluated across the entire dataset. In other words, we only provide learning curves for the datasets in which the confidence intervals of the best fixed policy overlap with those of the winning policy in at least one of the Tables \ref{tab:NB10_stationary_AAA}--\ref{tab:PERC10_stationary_AAA}.

\subsubsection{Experiments on non-stationary datasets.}
\begin{figure}\label{fig:NB10_nonstationary}%[H]
\centering
\subfigure{\includegraphics[width=0.485\textwidth]{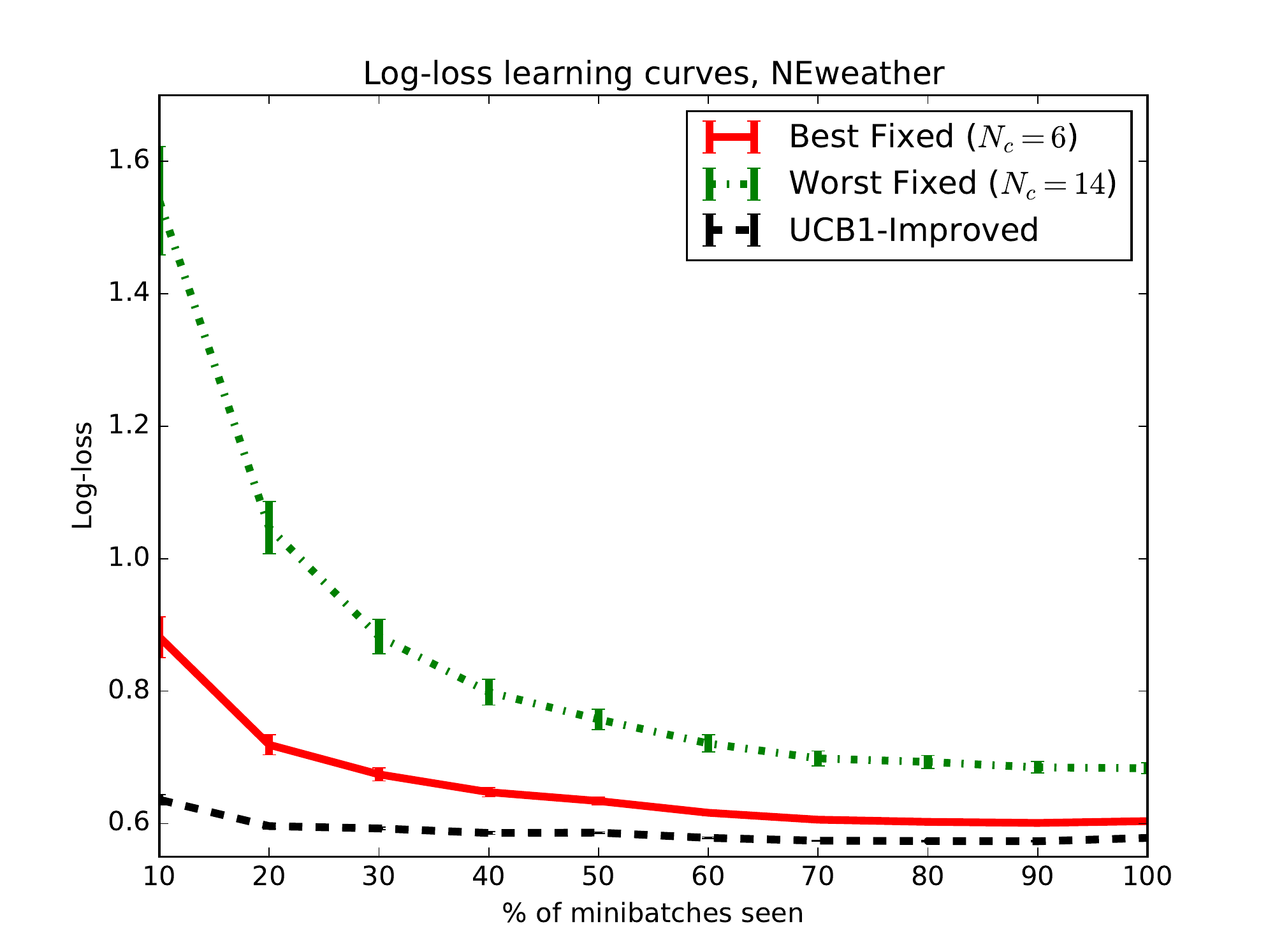}
  \label{}}
  %\vspace{-3.75cm}
%\bigskip
  %\vspace{-5cm}
\subfigure{\includegraphics[width=0.485\textwidth]{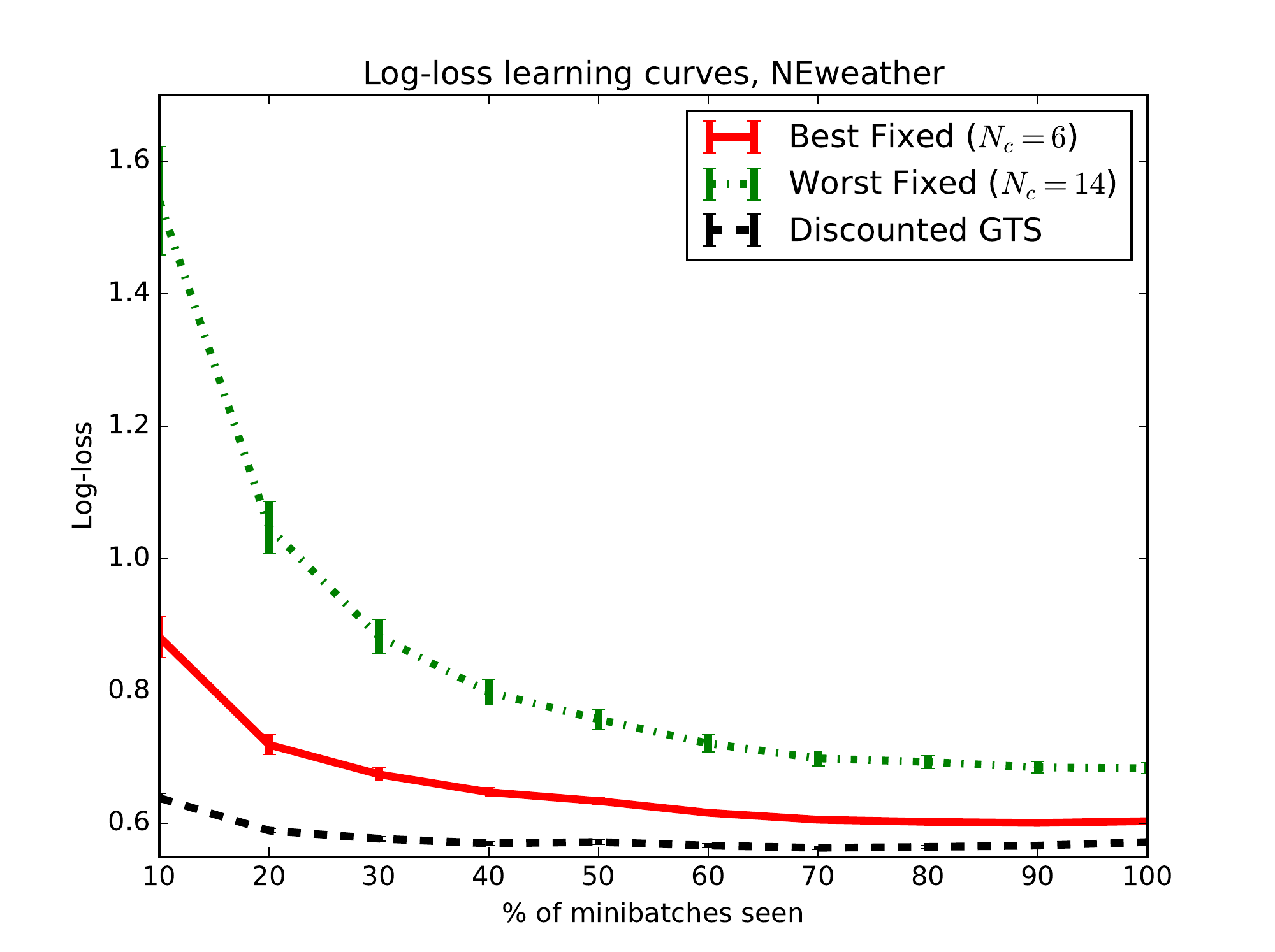}
  \label{}}
  %\vspace{-3.75cm}
%\bigskip
  %\vspace{-5cm}
\subfigure{\includegraphics[width=0.485\textwidth]{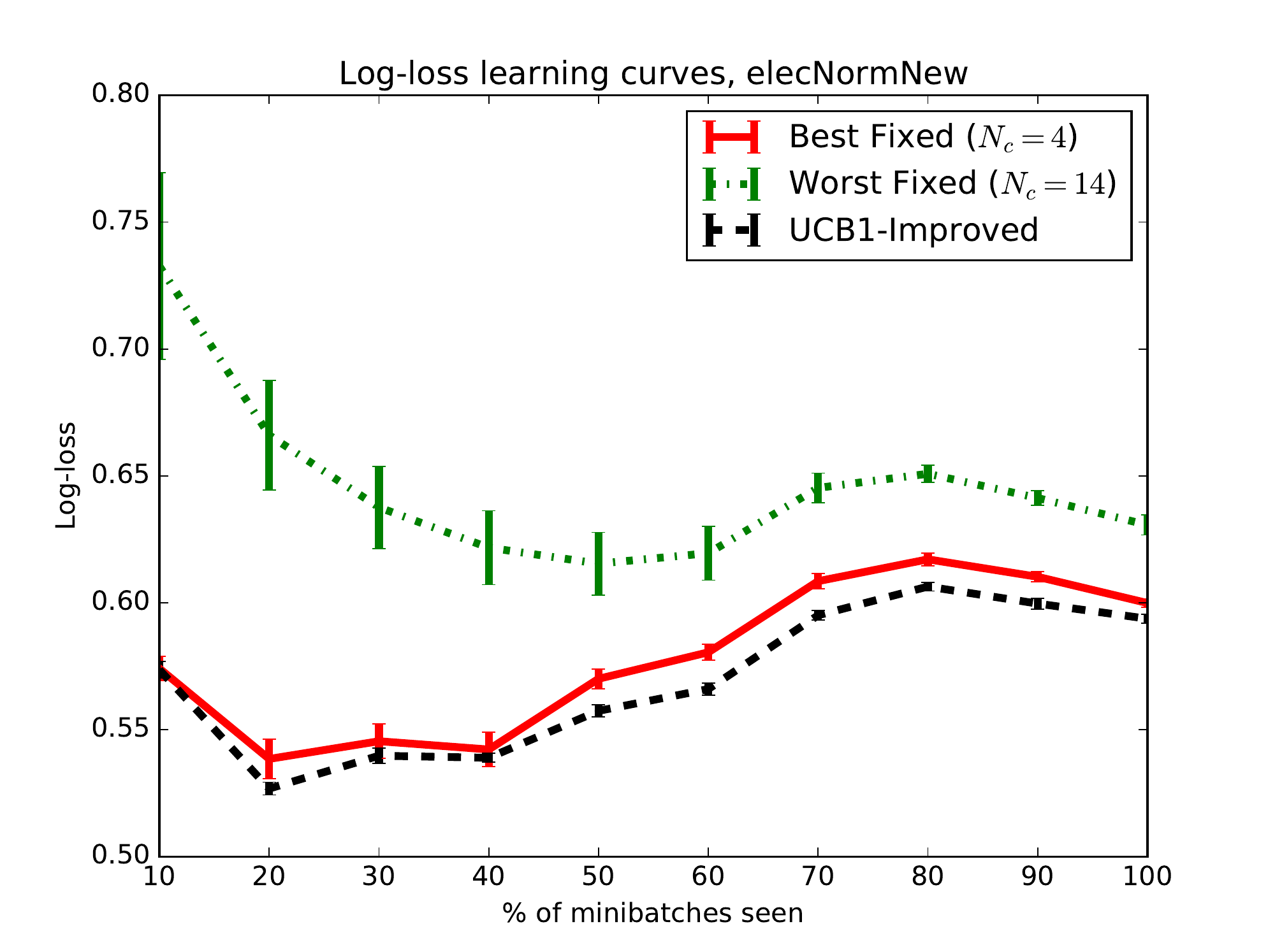}
  \label{}}
  %\vspace{-2.5cm}
%\bigskip
%\vspace{-1cm}
\subfigure{\includegraphics[width=0.485\textwidth]{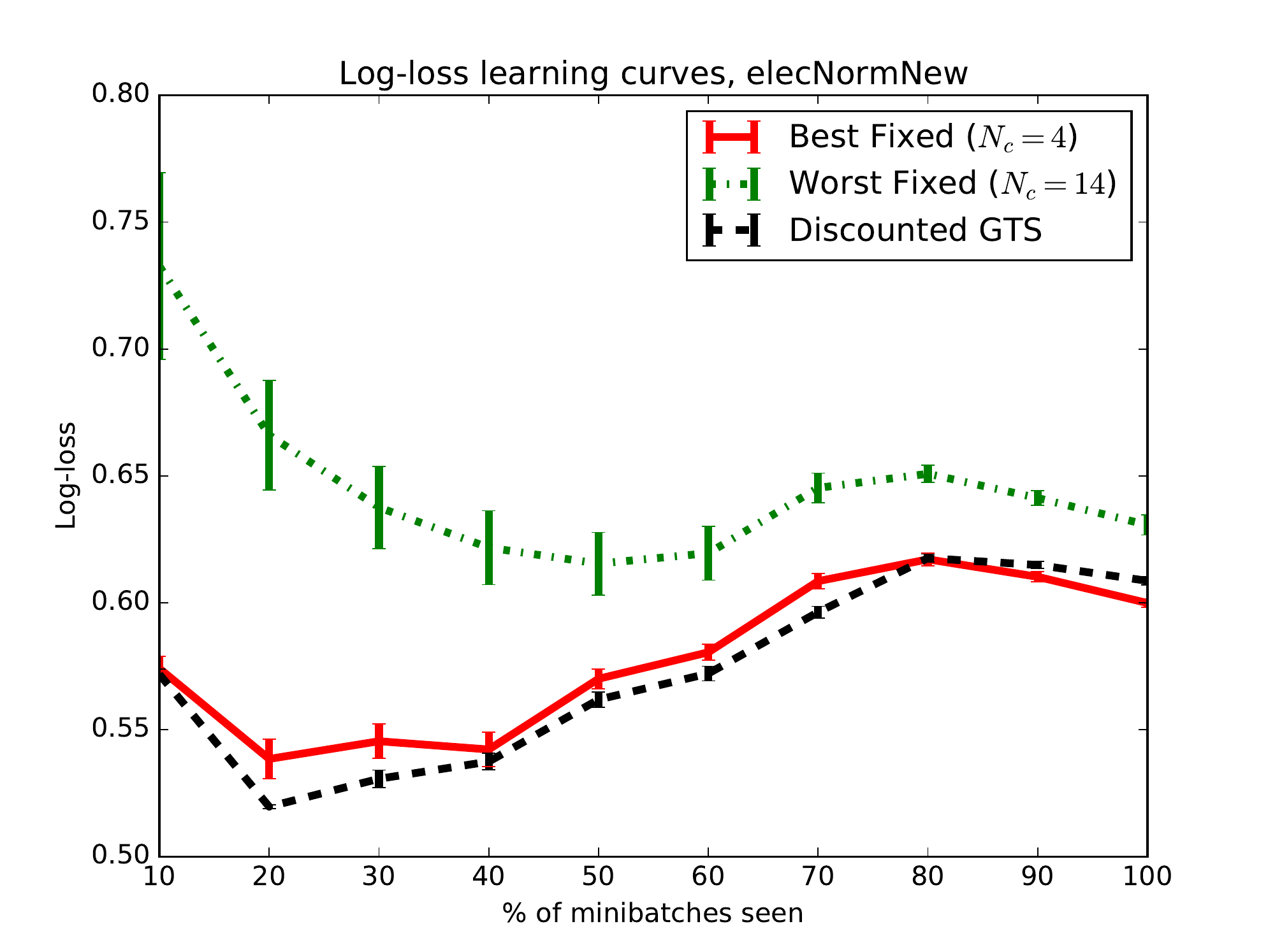}
  \label{}}
  %\vspace{-1cm}
  \subfigure{\includegraphics[width=0.485\textwidth]{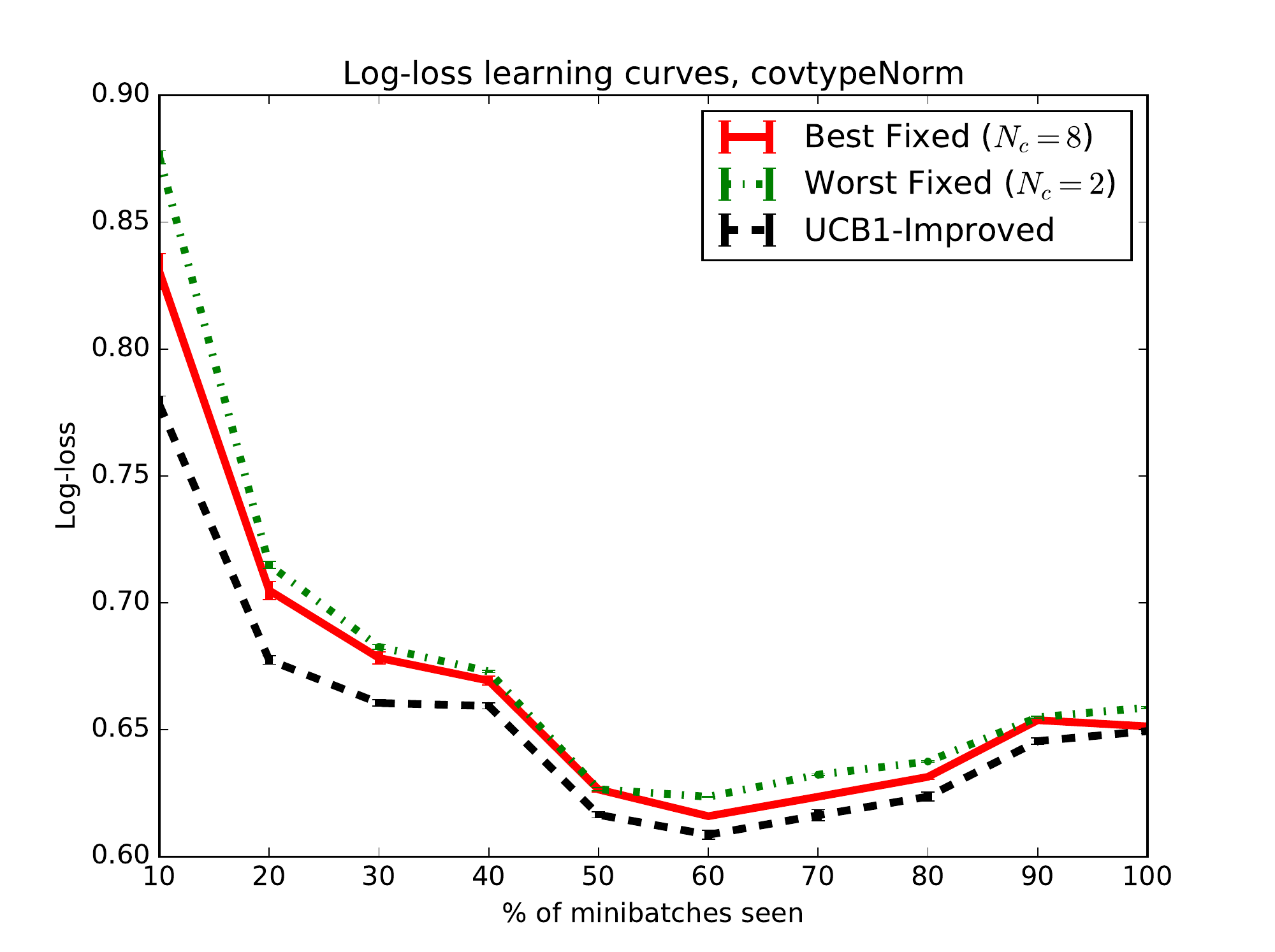}
  \label{}}
  %\vspace{-2.5cm}
%\bigskip
%\vspace{-1cm}
\subfigure{\includegraphics[width=0.485\textwidth]{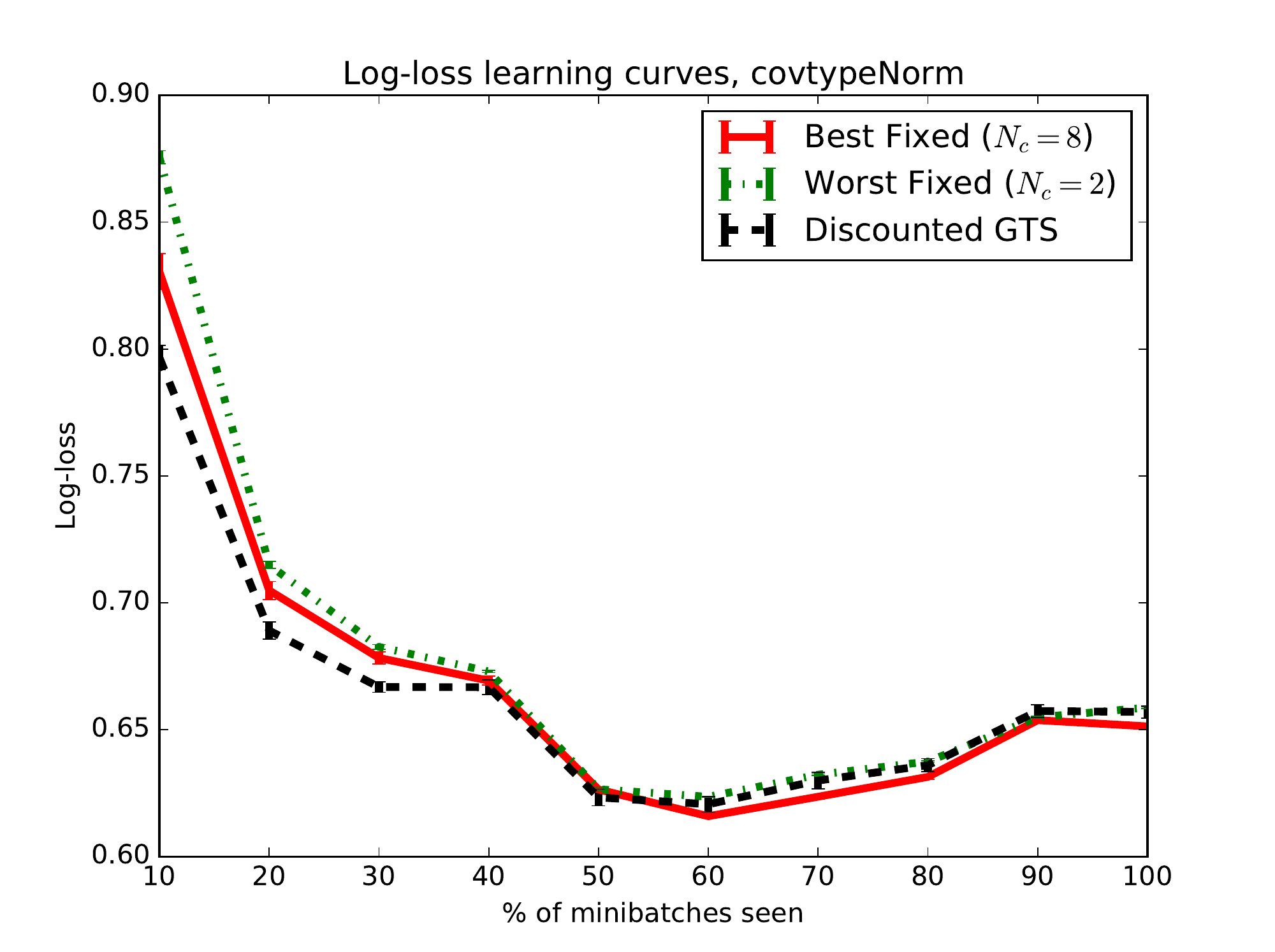}
  \label{}}
  %\vspace{-1cm}
  \caption{Log-loss versus number of minibatches seen. Lower values correspond to better probability estimation. The best and worst fixed calibration policies are compared against calibration under the non-discounted UCB1-Improved policy [LEFT] or the discounted Gaussian Thompson Sampling policy [RIGHT]. Results for Naive Bayes, $T=25$ on nonstationary datasets. Uncalibrated OnlineBoost curves were omitted due to their large log-loss (see Table \ref{tab:NB10_nonstationary}).}
\end{figure}

Next, we present experiments on the non-stationary datasets for Naive Bayes with $T=25$ in Table \ref{tab:NB10_nonstationary} and Figure \ref{fig:NB10_nonstationary}. The general pattern we observed in the previous set of experiments also appears here. Inspecting the largest dataset used in our study, \emph{forest}, we can see that the different calibration policies produce similar results with one another. This appears to be because ample datapoints are available and the feature space is relatively small (the dataset consists of $581,012$ datapoints and only $54$ features) to allow learning both good ensemble parameters and calibrator parameters regardless of the relative amount of data used for each of these learning tasks.

\begin{table}[H]
%\scriptsize
\center
\caption{Log-loss across the entire dataset. Lowest average value shown in bold. Results for Naive Bayes, $T=25$ on nonstationary datasets. The low variance is due to (i) not shuffling the datapoints on each run, respecting the order in which they appear \& (ii) the larger ratio of datapoints to features compared to the stationary datasets used.}\label{tab:NB10_nonstationary}
%\vspace{-0.2cm}
    \begin{tabular}{|c|c|c|c|c|c|c|c|c|c|}
    \hline
 	\multirow{3}{*}{Dataset} &   & Best & Worst &  &  UCB1 &   & Disc. & Disc. & Disc. \\
 	 & Uncalibrated & Fixed & Fixed & UCB1 & Improved & GTS & UCB1 & UCB1 & GTS\\ 
     &  &  &  &  &  &  &  & Improved & \\ \hline\hline    
        \multirow{3}{*}{\emph{weather}} & $1.599$ & $0.583$ & $0.683$ & $0.579$ & $0.578$ & $0.585$ & $0.760$ &  $0.739$ &  $\mathbf{0.572}$ \\
       & $\pm$ & $\pm$ & $\pm$ & $\pm$ & $\pm$ & $\pm$ & $\pm$ & $\pm$ & $\mathbf{\pm}$ \\
       & $0.062$ & $0.002$ & $0.008$ & $0.001$ & $0.001$ & $0.007$ & $0.083$ & $0.036$ & $\mathbf{0.002}$ \\ \hline              
    \multirow{3}{*}{\emph{electricity}} & $4.662$ & $0.604$ & $0.631$ & $0.599$ & $\mathbf{0.594}$ & $0.608$ & $0.612$ & $0.601$ & $0.609$ \\ 
     & $\pm$ & $\pm$ & $\pm$ & $\pm$ & $\mathbf{\pm}$ & $\pm$ & $\pm$ & $\pm$ & $\pm$ \\
     & $0.126$ & $0.001$ & $0.004$ & $0.001$ & $\mathbf{0.002}$ & $0.002$ & $0.002$ & $0.009$ & $0.002$ \\ \hline                      
    \multirow{3}{*}{\emph{forest}} & $5.618$ & $0.645$ & $0.660$ & $0.645$ & $\mathbf{0.643}$ & $0.647$ & $0.677$ & $0.648$ & $0.657$ \\ 
     & $\pm$ & $\pm$ & $\pm$ & $\pm$ & $\mathbf{\pm}$ & $\pm$ & $\pm$ & $\pm$ & $\pm$ \\
     & $0.010$ & $0.001$ & $0.001$ & $0.002$ & $\mathbf{0.001}$ & $0.001$ & $0.015$ & $0.006$ & $0.004$ \\ \hline              
    \end{tabular}
   % \vspace{-3cm}
\end{table}
%\vspace{-1.25cm}

\subsection{General observations}
Overall, calibration under any policy almost always improves the probability estimation w.r.t. uncalibrated OnlineBoost. As far as fixed policies are concerned, the policy with $N_c = 2$ and that with $N_c = 4$ dominated the rest in our experiments, while $N_c = 12$ and $N_c = 14$ were the values that led to the poorest probability estimates. This was to be expected, as it suggests that the more frequently we calibrate, the better the resulting probability estimates will be (averaged across all rounds).

Moreover, we saw that certain bandit policies (UCB1 policies without reward discounting and discounted-reward Gaussian Thompson Sampling) consistently exceed or at least match the probability estimation performance of the best fixed calibration policy. These results are robust to the choice of weak learner and the presence or absence of non-stationarity in the dataset but also to the ensemble size and degree of weak learner regularization (see Appendix $C$).

It should also be noted that the computational cost of the bandit policies is the same as that of the fixed ones (assuming a given value of $N_c$ for the latter) and that finding the best fixed policy requires a search over the possible values of $N_c$, something impossible in an online setting without parallelization or increasing the computational cost. Finally, bandit policies allow the ratio of training over calibrating steps to be adaptive, unlike, fixed policies. All these reasons make bandit policies superior to naive calibration.

Overall, discounting rewards --at least as applied to our experiments-- considerably improves Thompson Sampling, but greatly deteriorates UCB-based policies. In Thompson Sampling, discounting increases the variance of the reward posteriors. This increases the probability of the currently non-optimal arm being pulled. On the other hand, we observed that discounted UCB policies tended to get `trapped' to situations where only one action (either `train' or `calibrate') was taken. In discounted UCB policies, the padding function's value for an action shrinks with the number of times it has been performed. It appears that in the situations examined, the padding function's value for the action performed shrinks more and more slowly and never gets to the point where it eventually allows the upper confidence bound of the other action to overtake its own. 

\section{Conclusion and Future Work}

We examined probability estimation in online boosting ensembles and found that the scores they generate are distorted in a systematic fashion. We saw that --as in the case of batch boosting-- calibration can greatly improve the probability estimates. We resolved the problem of deciding when to train the ensemble and when to calibrate with the use of bandit optimization. More specifically, UCB1 policies\cite{Auer2002,garivier43,KaufmannCG12} without reward discounting and Thompson Sampling\cite{thompson1933,agrawal2011,ChapelleL11}, especially with reward discounting were found to perform at least as well as the best naive calibration policy in terms of probability estimation in every experiment. 

The merits of using bandit policies over naive calibration are manifold. Not only is the overall probability estimation performance superior, but it also converges much faster than the latter (in terms of minibatches seen). Moreover, to find the best naive calibration policy (i.e. fixed policy of calibrating on every $N_c$ rounds), we need to either determine the value of $N_c$ in advance, which is not possible in an online --possibly non-stationary-- setting. Furthermore, a fixed policy would be unable to adapt to non-stationarity: the optimal ratio of train and calibration actions might need to change during the course of training. The memory and computational complexity of the Bandit policies we examine here is constant w.r.t. the number of examples seen (as is required for online learning) and is the same as that of the fixed policies. We found that the superiority of these policies is robust to the choice of weak learner, ensemble size, degree of regularization and across datasets --both stationary and non-stationary. Finally, it would be straightforward to apply the same techniques to other types of learning tasks e.g. cost-sensitive or imbalanced class learning (situations in which good probability estimates are necessary for making decisions), simply by changing the reward function. 

%In fact the execution time for any form of calibrated boosting is expected to be smaller than its uncalibrated couneterpart as training the ensemble is typically slower than training the calibration

The purpose of this work was not to determine the best calibration method for an online boosting ensemble, but rather to identify fast, flexible and successful policies for balancing between training the ensemble and the calibrator (regardless of their specifics) in a hyperparameter-free fashion. In future work, to improve probability estimation, we can use the online version of isotonic regression~\cite{kotlowski2016online}, or produce online adaptations of spline calibration~\cite{goldstein2015} or beta calibration~\cite{kull2017} to calibrate the scores of the ensemble. Isotonic regression is a non-parametric method that can capture score distortions of any non-decreasing form. It is prone to overfitting in the presence of small samples, but is expected to outperform logistic calibration as the number of available data grows. Spline calibration is a smoothed version of the --piecewise-linear-- former. Beta calibration is an improvement over logistic calibration, especially when score distributions are heavily skewed or when the scores happen to already be calibrated. %Another direction not explored is the optimization function used to train the calibrator.

An alternative direction for future work could be to explore different bandit policies based on more relaxed learning assumptions. These could include the use of contextual bandits~\cite{krause2011contextual} that take into account the feature vector ${\bf{x}}$ of new instances when selecting the next action. Another family of bandit policies worth exploring is that of adversarial bandits~\cite{Auer2003}, which can handle environments that adapt to our actions. 

\bibliography{BetterBoostingBandits}

\newpage

\section*{Appendix A: Proof sketch of Theorem 1}

\begin{theorem*}
OnlineBoost greedily minimizes the exponential loss of the margin $L(y,F(\mathbf{x})) = e^{-yF(\mathbf{x})}$ via stochastic gradient descent steps in the space of functions $F(\mathbf{x})$.
\end{theorem*}

\begin{proof_sketch} 
In Algorithm \ref{alg:OnlineBoost_A}, we repeat the OnlineBoost algorithm for convenience.

\begin{algorithm}[htb]
   \caption{OnlineBoost (Original version)}
   \label{alg:OnlineBoost_A}
\begin{algorithmic}
\STATE {\bfseries Input}: Number of weak learners $T$, training examples $\{(\mathbf{x}_i, y_i)|i =1, \dots, N\}$\\~~~~~~~~~~presented one at a time
\STATE  {\bfseries For each $i$ do}:
\STATE  {} \ \ \ \ \ Set example weight $\lambda = 1$
\STATE  \ \ \ \ \ {\bfseries For each $t \in {1, 2, \dots, T}$ do}:
\STATE {} \ \ \ \ \ \ \ \ \ \ Set $k$ according to $Poisson(\lambda)$
\STATE {} \ \ \ \ \ \ \ \ \ \ {\bfseries Do $k$ times}:
\STATE {} \ \ \ \ \ \ \ \ \ \ \ \ \ \ \ $h_t \leftarrow OnlineLearnAlg(h_t,  (\mathbf{x}_i, y_i))$
\STATE {} \ \ \ \ \ \ \ \ \ \ {\bfseries If $h_t(\mathbf{x}_i)= y_i$}:
\STATE {} \ \ \ \ \ \ \ \ \ \ \ \ \ \ \ $\lambda^{sc}_t \leftarrow \lambda^{sc}_t +\lambda$
\STATE {} \ \ \ \ \ \ \ \ \ \ \ \ \ \ \ $\epsilon_t = \frac{\lambda^{sw}_t}{\lambda^{sw}_t + \lambda^{sc}_t}$
\STATE {} \ \ \ \ \ \ \ \ \ \ \ \ \ \ \ $\lambda \leftarrow \lambda \times \frac{1}{2(1-\epsilon_t)}$
\STATE {} \ \ \ \ \ \ \ \ \ \ {\bfseries Else}: 
\STATE {} \ \ \ \ \ \ \ \ \ \ \ \ \ \ \ $\lambda^{sw}_t \leftarrow \lambda^{sw}_t +\lambda$
\STATE {} \ \ \ \ \ \ \ \ \ \ \ \ \ \ \ $\epsilon_t = \frac{\lambda^{sw}_t}{\lambda^{sw}_t + \lambda^{sc}_t}$
\STATE {} \ \ \ \ \ \ \ \ \ \ \ \ \ \ \ $\lambda \leftarrow  \lambda \times \frac{1}{2\epsilon_t}$
\\\hrulefill
\STATE {\bfseries Prediction}: On example $(\mathbf{x}, y)$, predict $H(\mathbf{x}) = sign \Big[ \sum_{t=1}^{T}{h_t(\mathbf{x})\log{\frac{1-\epsilon_t}{\epsilon_t}}} \Big]$
\end{algorithmic}
\end{algorithm}

In Algorithm \ref{alg:OnlineBoost_2_A} we rewrite OnlineBoost in such a way that the weight assigned to each example $i \in \{1, \dots, N\}$ for the purposes of updating the parameters of each base learner $t \in \{1, \dots, T\}$ is stored in a separate variable $\lambda_t(i)$, rather than overwritten on each update on the variable $\lambda$. Storing so many variables is --of course-- prohibitive in the online setting, but here we do this merely for illustrative purposes.

The equivalence of Algorithm \ref{alg:OnlineBoost_2_A} to Algorithm \ref{alg:OnlineBoost_A} is trivial, thus we will not give an explicit proof. It is based on the fact that $\lambda$ in Algorithm \ref{alg:OnlineBoost_A} will equal to $\lambda_{t-1}(i)$ of Algorithm \ref{alg:OnlineBoost_2_A} before updating the weights of the $t$-th weak learner according to the $i$-th datapoint and to $\lambda_{t}(i)$ after. Once all $T$ weak learners are trained on the $i$-th datapoint, $\lambda$ will become equal to $\lambda_{t}(i)$ and so on.

\begin{algorithm}[htb]
   \caption{OnlineBoost (storing all weights)}
   \label{alg:OnlineBoost_2_A}
\begin{algorithmic}
\STATE {\bfseries Input}: Number of weak learners $T$, training examples $\{(\mathbf{x}_i, y_i)|i = 1, \dots, N\}$ available one at a time
\STATE  {\bfseries For each $i$ do}:
\STATE  {} \ \ \ \ \ Set example weight $\lambda_0(i) = 1$
\STATE  \ \ \ \ \ {\bfseries For each $t \in \{1, 2, \dots, T\}$ do:}
\STATE {} \ \ \ \ \ \ \ \ \ \ Set $k$ according to $Poisson(\lambda_{t-1}(i))$
\STATE {} \ \ \ \ \ \ \ \ \ \ {\bfseries Do $k$ times}:
\STATE {} \ \ \ \ \ \ \ \ \ \ \ \ \ \ \ $h_t \leftarrow OnlineLearnAlg(h_t,  (\mathbf{x}_i, y_i))$
\STATE {} \ \ \ \ \ \ \ \ \ \ {\bfseries If $h_t(\mathbf{x}_i)= y_i$}:
\STATE {} \ \ \ \ \ \ \ \ \ \ \ \ \ \ \ $\lambda^{sc}_t \leftarrow \lambda^{sc}_t +\lambda_{t-1}(i)$
\STATE {} \ \ \ \ \ \ \ \ \ \ \ \ \ \ \ $\epsilon_t = \frac{\lambda^{sw}_t}{\lambda^{sw}_t + \lambda^{sc}_t}$
\STATE {} \ \ \ \ \ \ \ \ \ \ \ \ \ \ \ $\lambda_{t}(i) \leftarrow \lambda_{t-1}(i) \times \frac{1}{2(1-\epsilon_t)}$
\STATE {} \ \ \ \ \ \ \ \ \ \ {\bfseries Else}:
\STATE {} \ \ \ \ \ \ \ \ \ \ \ \ \ \ \ $\lambda^{sw}_t \leftarrow \lambda^{sw}_t + \lambda_{t-1}(i)$
\STATE {} \ \ \ \ \ \ \ \ \ \ \ \ \ \ \ $\epsilon_t = \frac{\lambda^{sw}_t}{\lambda^{sw}_t + \lambda^{sc}_t}$
\STATE {} \ \ \ \ \ \ \ \ \ \ \ \ \ \ \ $\lambda_{t}(i) \leftarrow  \lambda_{t-1}(i) \times \frac{1}{2\epsilon_t}$
\\\hrulefill
\STATE {\bfseries Prediction}:  On example $(\mathbf{x}, y)$, predict $H(\mathbf{x}) = sign \Big[ \sum_{t=1}^{T}{h_t(\mathbf{x})\log{\frac{1-\epsilon_t}{\epsilon_t}}} \Big]$
\end{algorithmic}
\end{algorithm}

Using the notation introduced in Algorithm \ref{alg:OnlineBoost_2_A}, we can denote the sums of weights corresponding to correctly and incorrectly classified examples by the $t$-th weak learner so far, respectively, as
    
\begin{equation}
\label{eq:lambda_sc_sw_A}
\lambda^{sc}_t =  \sum_{\substack{j: h_t(\mathbf{x}_j) = y_j \\ j \leq i} }{\lambda_{t-1}(j)}
   \quad\text{and}\quad 
\lambda^{sw}_t =  \sum_{\substack{j: h_t(\mathbf{x}_j) \neq y_j \\ j \leq i} }{\lambda_{t-1}(j)}
\end{equation}

\begin{algorithm}[htb]
   \caption{OnlineBoost (Reformulated --resampling version)}
   \label{alg:OnlineBoost_3_A}
\begin{algorithmic}
\STATE {\bfseries Input}: Number of weak learners $T$, training examples $\{(\mathbf{x}_i, y_i)|i =1, \dots, N\}$ available one at a time
\STATE  {\bfseries For each $i$ do}:
\STATE  {} \ \ \ \ \ Set example weight $\lambda_0(i) = 1$
\STATE  \ \ \ \ \ {\bfseries For each $t \in {1, 2, \dots, T}$ do}:
\STATE {} \ \ \ \ \ \ \ \ \ \ Set $k$ according to $Poisson(\lambda_{t-1}(i))$
\STATE {} \ \ \ \ \ \ \ \ \ \ {\bfseries Do $k$ times}:  
\STATE {} \ \ \ \ \ \ \ \ \ \ \ \ \ \ \ $h_t \leftarrow OnlineLearnAlg(h_t,  (\mathbf{x}_i, y_i))$
\STATE {} \ \ \ \ \ \ \ \ \ \ $\epsilon_t = \frac{\sum_{\substack{j: h_t(\mathbf{x}_j) \neq y_j \\  j \leq i} }{\lambda_{t-1}(j)}} {\sum_{\substack{j: h_t(\mathbf{x}_j) \neq y_j \\  j \leq i} }{\lambda_{t-1}(j)} + \sum_{\substack{j: h_t(\mathbf{x}_j) = y_j \\  j \leq i} }{\lambda_{t-1}(j)}}$
\STATE {} \ \ \ \ \ \ \ \ \ \ $\beta_t = \frac{1}{2}\log{\frac{1-\epsilon_t}{\epsilon_t}}$
\STATE {} \ \ \ \ \ \ \ \ \ \ $\lambda_{t}(i) = \frac{1}{Z_t} \lambda_{t-1}(i) e^{-y_i \beta_t h_t(\mathbf{x}_i)}$, where $Z_t = \frac{2(1-\epsilon_t)}{e^{\beta_t}}$ (constant w.r.t. $i$)
\\\hrulefill
\STATE {\bfseries Prediction:} On example $(\mathbf{x}, y)$, predict $H(\mathbf{x}) = sign \Big[ \sum_{t=1}^{T}{\beta_t h_t(\mathbf{x})} \Big]$
\end{algorithmic}
\end{algorithm}

We now claim that Algorithm \ref{alg:OnlineBoost_3_A} is another reformulation of OnlineBoost, thus  equivalent to Algorithm \ref{alg:OnlineBoost_2_A}. This is less straightforward to see, so we shall prove that the steps of the two algorithms are equivalent. Notice that the changes w.r.t. Algorithm \ref{alg:OnlineBoost_2_A} occur in the weight update rule, the calculation of the confidence coefficients, and the final decision rule. Let us inspect these to verify that the two algorithms are equivalent.

\paragraph*{Confidence coefficients:} Algorithm \ref{alg:OnlineBoost_2_A}, does not explicitly store the confidence coefficients (i.e. the voting weights $\beta_t$) of the weak learners. It does not need to, as $\beta_t$ is calculated based on $\epsilon_t$, which in turn only requires $\lambda^{sc}_t$ and $\lambda^{sw}_t$. However, as we will see, in Algorithm \ref{alg:OnlineBoost_3_A}, we can define the quantity,
\begin{equation}
\label{eq:beta_A}
\beta_t = \frac{1}{2}\log{\frac{1-\epsilon_t}{\epsilon_t}},
\end{equation}
where --as we will see-- $\epsilon_t$ is the same in both algorithms, and use it to produce equivalent weight update and prediction rules.  

\paragraph*{Weight updates:}
In Algorithm \ref{alg:OnlineBoost_3_A}, the weight of the $i$-th example for the purposes of updating the parameters of the next weak learner is given by the equation
\begin{equation}
\label{eq:lambda_t_i_A}
\lambda_{t}(i) = \frac{1}{Z_t} \lambda_{t-1}(i) e^{-y_i \beta_t h_t(\mathbf{x}_i)},
   \quad\text{where}\quad 
Z_t = \frac{2(1-\epsilon_t)}{e^{\beta_t}}
\end{equation}
is used for normalization. This weight update rule can be rewritten as
\begin{equation*}
\begin{split}
\lambda_{t}(i) & \stackrel{\text{Eq.(\ref{eq:lambda_t_i_A})}}{=} \frac{e^{\beta_t}}{2(1-\epsilon_t)} \lambda_{t-1}(i) e^{-y_i \beta_t h_t(\mathbf{x}_i)} \stackrel{\text{Eq.(\ref{eq:beta_A})}}{=} \\
			   & = \frac{e^{\frac{1}{2}\log{\frac{1-\epsilon_t}{\epsilon_t}}}}{2(1-\epsilon_t)} \lambda_{t-1}(i) e^{-y_i \frac{1}{2}\log{\frac{1-\epsilon_t}{\epsilon_t}} h_t(\mathbf{x}_i)} = \\
			   & = \frac{e^{\log{\big(\frac{1-\epsilon_t}{\epsilon_t}\big)^{\frac{1}{2}}}}}{2(1-\epsilon_t)} \lambda_{t-1}(i) e^{\log{\big(\frac{1-\epsilon_t}{\epsilon_t}\big)}^{-\frac{1}{2}y_i h_t(\mathbf{x}_i)}} = \\
			   & = \lambda_{t-1}(i) \frac{\big(\frac{1-\epsilon_t}{\epsilon_t}\big)^{\frac{1}{2}} \big(\big(\frac{1-\epsilon_t}{\epsilon_t}\big)^{\frac{1}{2}}\big)^{-y_i h_t(\mathbf{x}_i)}}{2(1-\epsilon_t)} = \\
			   & = 
  \begin{cases}
    \lambda_{t-1}(i) \frac{1}{2(1-\epsilon_t)}, & h_t(\mathbf{x}_i) = y_i \\
    \lambda_{t-1}(i) \frac{1}{2\epsilon_t}~~~~~, & h_t(\mathbf{x}_i) \neq y_i.
  \end{cases}
\end{split}
\end{equation*}
And from Eq. (\ref{eq:lambda_sc_sw_A}), we have that
\begin{equation}
\label{eq:epsilon_A}
\epsilon_t = \frac{\sum_{\substack{j: h_t(\mathbf{x}_j) \neq y_j \\  j \leq i} }{\lambda_{t-1}(j)}} {\sum_{\substack{j: h_t(\mathbf{x}_j) \neq y_j \\  j \leq i} }{\lambda_{t-1}(j)} + \sum_{\substack{j: h_t(\mathbf{x}_j) = y_j \\  j \leq i} }{\lambda_{t-1}(j)}} = \frac{\lambda^{sw}_t}{\lambda^{sw}_t + \lambda^{sc}_t}.
\end{equation}
Therefore, the weight update rules of the two algorithms are identical.

%Finally, since $\lambda(i)$ of Algorithm \ref{alg:OnlineBoost_2_A} is equal to $\lambda_{t-1}(i)$ of Algorithm \ref{alg:OnlineBoost_3_A} before the $t$-th weight update and $\lambda_{t}(i)$ after it, \emph{the weight update rules of the two algorithms are identical}.

\paragraph*{Prediction Rule:}
In the new formulation of Algorithm \ref{alg:OnlineBoost_3_A}, the predictions are given by
\begin{equation*}
\begin{split}
H(x) &= sign \Big[ \sum_{t=1}^{T}{\beta_t h_t(\mathbf{x})} \Big] \stackrel{\text{Eq.(\ref{eq:beta_A})}}{=} sign \Big[ \sum_{t=1}^{T}{h_t(\mathbf{x})\frac{1}{2}\log{\frac{1-\epsilon_t}{\epsilon_t}}} \Big] =\\
     &= sign \Big[ \frac{1}{2}\sum_{t=1}^{T}{h_t(\mathbf{x})\log{\frac{1-\epsilon_t}{\epsilon_t}}} \Big] = sign \Big[ \sum_{t=1}^{T}{h_t(\mathbf{x})\log{\frac{1-\epsilon_t}{\epsilon_t}}} \Big].
\end{split}
\end{equation*}
So, from Eq. (\ref{eq:epsilon_A}), we have that Algorithm \ref{alg:OnlineBoost_2_A} and Algorithm \ref{alg:OnlineBoost_3_A} use the same prediction rule. Therefore all the steps of Algorithm \ref{alg:OnlineBoost_2_A} and Algorithm \ref{alg:OnlineBoost_3_A} are equivalent, and since Algorithm \ref{alg:OnlineBoost_2_A} is a reformulation of OnlineBoost, so is 
Algorithm \ref{alg:OnlineBoost_3_A}.

Algorithm \ref{alg:OnlineBoost_4_A} is a final reformulation of OnlineBoost. $WeightedOnlineLearnAlg(h_t, (\mathbf{x}_i, y_i), \lambda_{t-1}(i))$ is a weighted online learning algorithm updates the parameters of the weak learner $h_t$ on example $(\mathbf{x}_i, y_i)$ using a weight of $\lambda_{t-1}(i)$. The rationale is that instead of training the new classifier $k$ times on the $i$-th example, we can train it once with a weight $k$ (i.e. taking the reweighting rather than the resampling approach to boosting). Moreover, as each time $k \sim Poisson(\lambda_{t-1}(i))$, for the purposes of updating the parameters of the $t$-th weak learner on the $i$-th example, we have that $\lambda_{t-1}(i) = \mathbb{E}[k]$. So by training the $t$-th weak learner on the $i$-th example with a weight of $\lambda_{t-1}(i)$ we get --in expectation-- the same updates. This is the only change w.r.t. Algorithm \ref{alg:OnlineBoost_3_A}\footnote{This step of the proof is somewhat redundant as simply establishing the equivalence of OnlineBoost to AdaBoost-by-reweighting would have been sufficient for the purposes of this paper.}. 

\begin{algorithm}[htb]
   \caption{OnlineBoost (Reformulated --reweighting version using expected weights $\lambda_{t-1}(i)$)}
   \label{alg:OnlineBoost_4_A}
\begin{algorithmic}
\STATE {\bfseries Input}: Number of weak learners $T$, training examples $\{(\mathbf{x}_i, y_i)|i =1, \dots, N\}$ available one at a time
\STATE  {\bfseries For each $i$ do}:
\STATE  {} \ \ \ \ \ Set example weight $\lambda_0(i) = 1$
\STATE  \ \ \ \ \ {\bfseries For each $t \in {1, 2, \dots, T}$ do}:
\STATE {} \ \ \ \ \ \ \ \ \ \ $h_t \leftarrow WeightedOnlineLearnAlg(h_t,  (\mathbf{x}_i, y_i), \lambda_{t-1}(i))$
\STATE {} \ \ \ \ \ \ \ \ \ \ $\epsilon_t = \frac{\sum_{\substack{j: h_t(\mathbf{x}_j) \neq y_j \\  j \leq i} }{\lambda_{t-1}(j)}} {\sum_{\substack{j: h_t(\mathbf{x}_j) \neq y_j \\  j \leq i} }{\lambda_{t-1}(j)} + \sum_{\substack{j: h_t(\mathbf{x}_j) = y_j \\  j \leq i} }{\lambda_{t-1}(j)}}$
\STATE {} \ \ \ \ \ \ \ \ \ \ $\beta_t = \frac{1}{2}\log{\frac{1-\epsilon_t}{\epsilon_t}}$
\STATE {} \ \ \ \ \ \ \ \ \ \ $\lambda_{t}(i) = \frac{1}{Z_t} \lambda_{t-1}(i) e^{-y_i \beta_t h_t(\mathbf{x}_i)}$, where $Z_t = \frac{2(1-\epsilon_t)}{e^{\beta_t}}$ (constant w.r.t. $i$)
\\\hrulefill
\STATE {\bfseries Prediction}: On example $(\mathbf{x}, y)$, predict $H(\mathbf{x}) = sign \Big[ \sum_{t=1}^{T}{\beta_t h_t(\mathbf{x})} \Big]$
\end{algorithmic}
\end{algorithm}

Finally, in Algorithm \ref{alg:AdaBoostFixedT} we present (the reweighting version of) batch AdaBoost with a fixed ensemble size $T$. It uses a batch learning algorithm to train the weak learner $h_t$ on the full dataset using weights $\lambda_{t-1}(i)$, $\forall i$, denoted as $WeightedBatchLearnAlg(\{(\mathbf{x}_i, y_i)|, \forall i\}, \{\lambda_{t-1}(i)|, \forall i\})$. 

%Comparing Algorithm \ref{alg:OnlineBoost_3_A} (equivalent to OnlineBoost by resampling) to Algorithm \ref{alg:AdaBoostFixedT} (batch AdaBoost by reweighting, with fixed ensemble size $T$), we see there are only two differences: The first difference is that Algorithm \ref{alg:OnlineBoost_3_A} updates the $t$-th base learner by calling the (unweighted) online learning algorithm \emph{OnlineLearnAlg()} $k$ times on the $i$-th example (and similarly for all examples). On the other hand, Algorithm \ref{alg:AdaBoostFixedT} updates the $t$-th base learner by calling the weighted batch learning algorithm \emph{BatchLearnAlg()} once, using $\lambda_{t-1}(i)$ as the weight for the $i$-th example. These two 

\begin{algorithm}[htb]
   \caption{AdaBoost (Fixed ensemble size $T$)}
   \label{alg:AdaBoostFixedT}
\begin{algorithmic}
\STATE {\bfseries Input}: Number of weak learners $T$, training examples $\{(\mathbf{x}_i, y_i)|i =1, \dots, N\}$ 
\STATE  {}Set example weight $\lambda_0(i) = 1$, $\forall i$
\STATE  {\bfseries For each $t \in \{1, 2, \dots, T\}$ do}:
\STATE {} \ \ \ \ \ \ \ \ \ \ $h_t \leftarrow WeightedBatchLearnAlg(\{(\mathbf{x}_i, y_i)|, \forall i\}, \{\lambda_{t-1}(i)|, \forall i\})$
\STATE {} \ \ \ \ \ \ \ \ \ \ $\epsilon_t = \frac{\sum_{i: h_t(\mathbf{x}_i) \neq y_i }{\lambda_{t-1}(i)}} {\sum_{i: h_t(\mathbf{x}_i) \neq y_i }{\lambda_{t-1}(i)} + \sum_{i: h_t(\mathbf{x}_i) = y_i }{\lambda_{t-1}(i)}}$
\STATE {} \ \ \ \ \ \ \ \ \ \ $\beta_t = \frac{1}{2}\log{\frac{1-\epsilon_t}{\epsilon_t}}$
\STATE {} \ \ \ \ \ \ \ \ \ \ $\lambda_{t}(i) = \frac{1}{Z_t} \lambda_{t-1}(i) e^{-y_i \beta_t h_t(\mathbf{x}_i)}$, $\forall i$, where $Z_t$ is constant w.r.t. $i$ %, where $Z_t = \frac{2(1-\epsilon_t)}{e^{\alpha_t}}$ (constant w.r.t. $i$)
\\\hrulefill
\STATE {\bfseries Prediction}: On example $(\mathbf{x}, y)$, predict $H(\mathbf{x}) = sign \Big[ \sum_{t=1}^{T}{\beta_t h_t(\mathbf{x})} \Big]$
\end{algorithmic}
\end{algorithm}

Comparing Algorithm \ref{alg:AdaBoostFixedT} to Algorithm \ref{alg:OnlineBoost_4_A} (equivalent to OnlineBoost), we see that the only difference is that all parameters estimated in the latter are updated only on the examples seen so far, rather than on the entire dataset (something natural for the online setting).

Since the steps of AdaBoost can be derived as (batch) gradient descent on an exponential loss of the margin $L(y,F(\mathbf{x})) = e^{-yF(\mathbf{x})}$ in the space of functions $F(\mathbf{x})$ \cite{friedman2000greedy, Mason1999}, and since OnlineBoost uses the same steps, but updates all parameters based on one example at a time, we can conclude that OnlineBoost minimizes the same loss function as AdaBoost (i.e. the exponential loss), but taking stochastic gradient descent steps in the space of functions $F(\mathbf{x})$.
\end{proof_sketch}

\newpage

\section*{Appendix B: Datasets used}
\begin{table}[H]
\center
\caption{Characteristics of the datasets used in our experiments; number of instances used, number of features, number of classes and presence or not of dataset shift. All stationary datasets can be found in the UCI repository. The non-stationary datasets were taken from~\cite{Losing2016}.}\label{tab:datasets}
%\vspace{-0.2cm}
    \begin{tabular}{|c|c|c|c|c|c|}
    \hline
 	\multirow{2}{*}{Dataset} & \# & \# & \# & Considered \\
 	 & Instances & Features & Classes & Stationary \\ \hline\hline    
    \emph{landsat} & $1,252$ & $36$ & $6$ & Yes \\ \hline  
    \emph{splice} & $1,524$ & $60$ & $3$ & Yes \\ \hline    
    \emph{musk2} & $2,034$ & $166$ & $2$ & Yes \\ \hline 
    \emph{krvskp} & $3,054$ & $36$ & $2$ & Yes \\ \hline
    \emph{waveform} & $3,306$ & $40$ & $3$ & Yes \\ \hline    
    \emph{spambase} & $3,626$ & $57$ & $2$ & Yes \\ \hline
    \emph{mushroom} & $7,832$ & $21$ & $2$ & Yes \\ \hline
    \emph{weather} & $18,159$ & $8$ & $2$ & No \\ \hline    
    \emph{electricity} & $45,312$ & $8$ & $2$ & No \\ \hline
    \emph{forest} & $581,012$ & $54$ & $7$ & No \\ \hline    
    \end{tabular}
   % \vspace{-3cm}
\end{table}
%\vspace{-1.25cm}

\newpage

\section*{Appendix C: Additional results}
Here we provide tables of log-loss across all examples, for each method, for all the experiments mentioned in the paper, along with learning curves of log-loss versus number of minibatches seen.

\subsection*{Effect of weak learner choice}
We provide results on the stationary datasets for ensembles of size $T=10$ for four different types of weak learners: Gaussian Naive Bayes, logistic regression, linear SVM and perceptron.

\begin{table}[H]
\center
\caption{Log-loss across the entire dataset. Lowest average value shown in bold. Results for Naive Bayes, $T=10$ on stationary datasets.}\label{tab:NB10_stationary}
%\vspace{-0.2cm}
    \begin{tabular}{|c|c|c|c|c|c|c|c|c|c|}
    \hline
 	\multirow{3}{*}{Dataset} &   & Best & Worst &  &  UCB1 &   & Disc. & Disc. & Disc. \\
 	 & Uncalibrated & Fixed & Fixed & UCB1 & Improved & GTS & UCB1 & UCB1 & GTS\\ 
     &  &  &  &  &  &  &  & Improved & \\ \hline\hline    
        \multirow{3}{*}{\emph{landsat}} & $0.138$ & $0.081$ & $0.131$ & $0.072$ & $\mathbf{0.071}$ & $0.074$ & $0.129$ &  $0.123$ &  $0.073$ \\
       & $\pm$ & $\pm$ & $\pm$ & $\pm$ & $\mathbf{\pm}$ & $\pm$ & $\pm$ & $\pm$ & $\pm$ \\
       & $0.002$ & $0.002$ & $0.004$ & $0.001$ & $\mathbf{0.001}$ & $0.002$ & $0.010$ & $0.011$ & $0.002$ \\ \hline              
    \multirow{3}{*}{\emph{splice}} & $0.444$ & $0.280$ & $0.410$ & $0.234$ & $\mathbf{0.229}$ & $0.282$ & $0.353$ & $0.376$ & $0.232$ \\ 
     & $\pm$ & $\pm$ & $\pm$ & $\pm$ & $\mathbf{\pm}$ & $\pm$ & $\pm$ & $\pm$ & $\pm$ \\
     & $0.034$ & $0.014$ & $0.014$ & $0.007$ & $\mathbf{0.009}$ & $0.022$ & $0.024$ & $0.037$ & $0.008$ \\ \hline              
    \multirow{3}{*}{\emph{musk2}} & $0.723$ & $0.343$ & $0.427$ & $\mathbf{0.320}$ & $0.332$ & $0.397$ & $0.711$ & $0.441$ & $0.345$ \\ 
     & $\pm$ & $\pm$ & $\pm$ & $\pm$ & $\mathbf{\pm}$ & $\pm$ & $\pm$ & $\pm$ & $\pm$ \\
     & $0.027$ & $0.008$ & $0.013$ & $\mathbf{0.006}$ & $0.003$ & $0.008$ & $0.025$ & $0.049$ & $0.008$ \\ \hline              
    \multirow{3}{*}{\emph{krvskp}} & $1.252$ & $0.488$ & $0.720$ & $0.474$ & $0.472$ & $0.486$ & $0.627$ & $0.767$ & $\mathbf{0.468}$ \\
       & $\pm$ & $\pm$ & $\pm$ & $\pm$ & $\pm$ & $\pm$ & $\pm$ & $\pm$ & $\mathbf{\pm}$ \\
       & $0.154$ & $0.014$ & $0.075$ & $0.019$ & $0.012$ & $0.016$ & $0.114$ & $0.114$ & $\mathbf{0.020}$ \\ \hline    
    \multirow{3}{*}{\emph{waveform}} & $0.824$ & $0.361$ & $0.466$ & $0.335$ & $\mathbf{0.334}$ & $0.342$ & $0.458$ & $0.492$ & $0.344$\\
       & $\pm$ & $\pm$ & $\pm$ & $\pm$ & $\pm$ & $\pm$ & $\pm$ & $\pm$ & $\pm$ \\
       & $0.012$ & $0.005$ & $0.009$ & $0.002$ & $\mathbf{0.004}$ & $0.007$ & $0.015$ & $0.024$ & $0.006$ \\ \hline              
    \multirow{3}{*}{\emph{spambase}} & $2.37$ & $0.532$ & $0.735$ & $0.493$ & $0.483$ & $\mathbf{0.481}$ & $0.536$ & $0.540$ & $0.489$\\
       & $\pm$ & $\pm$ & $\pm$ & $\pm$ & $\pm$ & $\mathbf{\pm}$ & $\pm$ & $\pm$ & $\pm$ \\
       & $0.026$ & $0.006$ & $0.017$ & $0.006$ & $0.005$ & $\mathbf{0.008}$ & $0.028$ & $0.037$ & $0.005$ \\ \hline                
    \multirow{3}{*}{\emph{mushroom}} & $0.770$ & $0.358$ & $0.503$ & $0.375$ & $0.335$ & $\mathbf{0.318}$ & $0.570$ & $0.570$ & $0.327$ \\ 
           & $\pm$ & $\pm$ & $\pm$ & $\pm$ & $\pm$ & $\pm$ & $\mathbf{\pm}$ & $\pm$ & $\pm$ \\
       & $0.057$ & $0.011$ & $0.017$ & $0.018$ & $0.020$ & $\mathbf{0.006}$ & $0.036$ & $0.072$ & $0.010$ \\ \hline  
    \end{tabular}
   % \vspace{-3cm}
\end{table}
%\vspace{-1.25cm}

\begin{table}[H]
\center
\caption{Log-loss across the entire dataset. Lowest average value shown in bold. Results for logistic regression, $T=10$ on stationary datasets. }\label{tab:LR10_stationary}
%\vspace{-0.2cm}
    \begin{tabular}{|c|c|c|c|c|c|c|c|c|c|}
    \hline
 	\multirow{3}{*}{Dataset} &   & Best & Worst &  &  UCB1 &   & Disc. & Disc. & Disc. \\
 	 & Uncalibrated & Fixed & Fixed & UCB1 & Improved & GTS & UCB1 & UCB1 & GTS\\ 
     &  &  &  &  &  &  &  & Improved & \\ \hline\hline    
        \multirow{3}{*}{\emph{landsat}} & $0.330$ & $\mathbf{0.106}$ & $0.158$ & $0.107$ & $0.108$ & $0.108$ & $0.173$ &  $0.203$ &  $0.127$ \\
       & $\pm$ & $\mathbf{\pm}$ & $\pm$ & $\pm$ & $\pm$ & $\pm$ & $\pm$ & $\pm$ & $\pm$ \\
       & $0.011$ & $\mathbf{0.002}$ & $0.004$ & $0.003$ & $0.004$ & $0.004$ & $0.019$ & $0.018$ & $0.013$ \\ \hline     
    \multirow{3}{*}{\emph{splice}} & $1.779$ & $0.626$ & $0.895$ & $\mathbf{0.506}$ & $0.554$ & $0.636$ & $0.711$ & $1.155$ & $0.584$ \\ 
     & $\pm$ & $\pm$ & $\pm$ & $\mathbf{\pm}$ & $\pm$ & $\pm$ & $\pm$ & $\pm$ & $\pm$ \\
     & $0.059$ & $0.023$ & $0.034$ & $\mathbf{0.029}$ & $0.020$ & $0.068$ & $0.117$ & $0.169$ & $0.034$ \\ \hline    
    \multirow{3}{*}{\emph{musk2}} & $1.276$ & $0.331$ & $0.398$ & $0.321$ & $0.330$ & $\mathbf{0.318}$ & $0.536$ & $0.514$ & $0.322$ \\ 
    & $\pm$ & $\pm$ & $\pm$ & $\pm$ & $\pm$ & $\mathbf{\pm}$ & $\pm$ & $\pm$ & $\pm$ \\
     & $0.015$ & $0.003$ & $0.008$ & $0.004$ & $0.004$ & $\mathbf{0.003}$ & $0.049$ & $0.038$ & $0.006$ \\ \hline       
    \multirow{3}{*}{\emph{krvskp}} & $1.151$ & $0.633$ & $0.881$ & $\mathbf{0.519}$ & $0.528$ & $0.645$ & $0.825$ & $0.733$ & $0.555$ \\
       & $\pm$ & $\pm$ & $\pm$ & $\mathbf{\pm}$ & $\pm$ & $\pm$ & $\pm$ & $\pm$ & $\pm$ \\
       & $0.073$ & $0.041$ & $0.042$ & $\mathbf{0.037}$ & $0.017$ & $0.048$ & $0.074$ & $0.041$ & $0.033$ \\ \hline 
    \multirow{3}{*}{\emph{waveform}} & $1.700$ & $0.444$ & $0.596$ & $\mathbf{0.427}$ & $0.429$ & $0.428$ & $0.503$ & $0.519$ & $0.464$\\
       & $\pm$ & $\pm$ & $\pm$ & $\pm$ & $\pm$ & $\pm$ & $\pm$ & $\pm$ & $\pm$ \\
       & $0.037$ & $0.004$ & $0.019$ & $\mathbf{0.003}$ & $0.007$ & $0.006$ & $0.026$ & $0.030$ & $0.016$ \\ \hline  
    \multirow{3}{*}{\emph{spambase}} & $1.182$ & $0.409$ & $0.572$ & $0.390$ & $0.395$ & $0.407$ & $0.540$ & $0.477$ & $\mathbf{0.389}$\\
       & $\pm$ & $\pm$ & $\pm$ & $\pm$ & $\pm$ & $\pm$ & $\pm$ & $\pm$ & $\mathbf{\pm}$ \\
       & $0.018$ & $0.006$ & $0.010$ & $0.005$ & $0.006$ & $0.007$ & $0.029$ & $0.024$ & $\mathbf{0.006}$ \\ \hline      
    \multirow{3}{*}{\emph{mushroom}} & $1.001$ & $0.343$ & $0.454$ & $0.322$ & $0.315$ & $\mathbf{0.312}$ & $0.531$ & $0.592$ & $0.330$ \\ 
           & $\pm$ & $\pm$ & $\pm$ & $\pm$ & $\pm$ & $\mathbf{\pm}$ & $\pm$ & $\pm$ & $\pm$ \\
       & $0.023$ & $0.007$ & $0.017$ & $0.007$ & $0.006$ & $\mathbf{0.007}$ & $0.026$ & $0.026$ & $0.006$ \\ \hline    
    \end{tabular}
   % \vspace{-3cm}
\end{table}
%\vspace{-1.25cm}

\begin{table}[H]
\center
\caption{Log-loss across the entire dataset. Lowest average value shown in bold. Results for linear SVM, $T=10$ on stationary datasets. }\label{tab:LINSVM10_stationary}
%\vspace{-0.2cm}
    \begin{tabular}{|c|c|c|c|c|c|c|c|c|c|}
    \hline
 	\multirow{3}{*}{Dataset} &   & Best & Worst &  &  UCB1 &   & Disc. & Disc. & Disc. \\
 	 & Uncalibrated & Fixed & Fixed & UCB1 & Improved & GTS & UCB1 & UCB1 & GTS\\ 
     &  &  &  &  &  &  &  & Improved & \\ \hline\hline    
        \multirow{3}{*}{\emph{landsat}} & $0.245$ & $ 0.130$ & $0.184$ & $0.118$ & $0.121$ & $0.127$ & $0.250$ & $0.243$ & $\mathbf{0.111}$ \\ 
           & $\pm$ & $\pm$ & $\pm$ & $\pm$ & $\pm$ & $\pm$ & $\pm$ & $\pm$ & $\mathbf{\pm}$ \\
       & $0.019$ & $0.009$ & $0.010$ & $0.008$ & $0.008$ & $0.010$ & $0.025$ & $0.034$ & $\mathbf{0.004}$ \\ \hline 
    \multirow{3}{*}{\emph{splice}} & $1.391$ & $0.663$ & $0.922$ & $0.622$ & $\mathbf{0.586}$ & $0.609$ & $0.816$ & $0.666$ & $0.640$ \\ 
           & $\pm$ & $\pm$ & $\pm$ & $\pm$ & $\mathbf{\pm}$ & $\pm$ & $\pm$ & $\pm$ & $\pm$ \\
       & $0.064$ & $0.018$ & $0.040$ & $0.031$ & $\mathbf{0.022}$ & $0.020$ & $0.105$ & $0.088$ & $0.038$ \\ \hline 
    \multirow{3}{*}{\emph{musk2}} & $0.953$ & $\mathbf{0.385}$ & $0.432$ & $0.400$ & $0.390$ & $0.406$ & $0.639$ & $0.622$ & $0.386$ \\ 
          & $\mathbf{\pm}$ & $\pm$ & $\pm$ & $\pm$ & $\pm$ & $\pm$ & $\pm$ & $\pm$ & $\pm$ \\
       & $\mathbf{0.020}$ & $0.008$ & $0.007$ & $0.006$ & $0.007$ & $0.008$ & $0.061$ & $0.066$ & $\mathbf{0.005}$ \\ \hline          
    \multirow{3}{*}{\emph{krvskp}} & $1.081$ & $0.671$ & $0.897$ & $\mathbf{0.608}$ & $0.648$ & $0.677$ & $0.706$ & $0.811$ & $0.667$ \\ 
           & $\pm$ & $\pm$ & $\pm$ & $\mathbf{\pm}$ & $\pm$ & $\pm$ & $\pm$ & $\pm$ & $\pm$ \\
       & $0.050$ & $0.022$ & $0.027$ & $\mathbf{0.021}$ & $0.027$ & $0.029$ & $0.047$ & $0.041$ & $0.030$ \\ \hline 
    \multirow{3}{*}{\emph{waveform}} & $1.156$ & $0.487$ & $0.615$ & $0.453$ & $\mathbf{0.451}$ & $\mathbf{0.451}$ & $0.611$ & $0.603$ & $0.464$ \\ 
          & $\pm$ & $\pm$ & $\pm$ & $\pm$ & $\mathbf{\pm}$ & $\mathbf{\pm}$ & $\pm$ & $\pm$ & $\pm$ \\
       & $0.045$ & $0.013$ & $0.015$ & $0.005$ & $\mathbf{0.004}$ & $\mathbf{0.005}$ & $0.021$ & $0.076$ & $0.005$ \\ \hline 
    \multirow{3}{*}{\emph{spambase}} & $0.893$ & $0.440$ & $0.551$ & $0.411$ & $\mathbf{0.410}$ & $0.418$ & $0.560$ & $0.612$ & $0.414$ \\ 
          & $\pm$ & $\pm$ & $\pm$ & $\pm$ & $\pm$ & $\pm$ & $\pm$ & $\pm$ & $\pm$ \\
       & $0.019$ & $0.008$ & $0.010$ & $0.005$ & $\mathbf{0.007}$ & $0.009$ & $0.028$ & $0.053$ & $0.0037$ \\ \hline  
    \multirow{3}{*}{\emph{mushroom}} & $0.673$ & $0.374$ & $0.489$ & $0.359$ & $0.363$ & $\mathbf{0.352}$ & $0.506$ & $0.573$ & $0.357$ \\ 
           & $\pm$ & $\pm$ & $\pm$ & $\pm$ & $\pm$ & $\mathbf{\pm}$ & $\pm$ & $\pm$ & $\pm$ \\
       & $0.015$ & $0.006$ & $0.015$ & $0.006$ & $0.006$ & $\mathbf{0.005}$ & $0.036$ & $0.039$ & $0.007$ \\ \hline     
    \end{tabular}
   % \vspace{-3cm}
\end{table}
%\vspace{-1.25cm}

%[TODO: FIGURES HERE]\\

\begin{table}[H]
\center
\caption{Log-loss across the entire dataset. Lowest average value shown in bold. Results for perceptron, $T=10$ on stationary datasets. }\label{tab:PERC10_stationary}
%\vspace{-0.2cm}
    \begin{tabular}{|c|c|c|c|c|c|c|c|c|c|}
    \hline
 	\multirow{3}{*}{Dataset} &   & Best & Worst &  &  UCB1 &   & Disc. & Disc. & Disc. \\
 	 & Uncalibrated & Fixed & Fixed & UCB1 & Improved & GTS & UCB1 & UCB1 & GTS\\ 
     &  &  &  &  &  &  &  & Improved & \\ \hline\hline    
        \multirow{3}{*}{\emph{landsat}} & $0.258$ & $ 0.120$ & $0.190$ & $0.125$ & $\mathbf{0.103}$ & $0.115$ & $0.164$ & $0.198$ & $0.120$ \\ 
           & $\pm$ & $\pm$ & $\pm$ & $\pm$ & $\pm$ & $\pm$ & $\pm$ & $\pm$ & $\pm$ \\
       & $0.012$ & $0.005$ & $0.009$ & $0.013$ & $\mathbf{0.004}$ & $0.005$ & $0.014$ & $0.028$ & $0.008$ \\ \hline      
    \multirow{3}{*}{\emph{splice}} & $1.453$ & $0.683$ & $0.879$ & $\mathbf{0.566}$ & $0.624$ & $0.642$ & $0.759$ & $0.651$ & $0.607$ \\ 
           & $\pm$ & $\pm$ & $\pm$ & $\mathbf{\pm}$ & $\pm$ & $\pm$ & $\pm$ & $\pm$ & $\pm$ \\
       & $0.037$ & $0.033$ & $0.050$ & $\mathbf{0.021}$ & $0.028$ & $0.025$ & $0.105$ & $0.022$ & $0.027$ \\ \hline        
    \multirow{3}{*}{\emph{musk2}} & $0.945$ & $0.392$ & $0.443$ & $0.390$ & $\mathbf{0.385}$ & $0.419$ & $0.591$ & $0.648$ & $0.408$ \\ 
          & $\pm$ & $\pm$ & $\pm$ & $\pm$ & $\mathbf{\pm}$ & $\pm$ & $\pm$ & $\pm$ & $\pm$ \\
       & $0.030$ & $0.006$ & $0.012$ & $0.006$ & $\mathbf{0.006}$ & $0.014$ & $0.044$ & $0.067$ & $0.013$ \\ \hline 
    \multirow{3}{*}{\emph{krvskp}} & $1.021$ & $0.745$ & $0.909$ & $0.632$ & $\mathbf{0.609}$ & $0.777$ & $0.812$ & $0.811$ & $0.611$ \\ 
           & $\pm$ & $\pm$ & $\pm$ & $\pm$ & $\mathbf{\pm}$ & $\pm$ & $\pm$ & $\pm$ & $\pm$ \\
       & $0.042$ & $0.031$ & $0.043$ & $0.037$ & $\mathbf{0.022}$ & $0.079$ & $0.069$ & $0.062$ & $0.017$ \\ \hline 
    \multirow{3}{*}{\emph{waveform}} & $1.106$ & $0.498$ & $0.589$ & $0.468$ & $0.492$ & $0.486$ & $0.573$ & $0.686$ & $\mathbf{0.456}$ \\ 
          & $\pm$ & $\pm$ & $\pm$ & $\pm$ & $\pm$ & $\pm$ & $\pm$ & $\pm$ & $\pm$ \\
       & $0.027$ & $0.010$ & $0.021$ & $0.010$ & $0.034$ & $0.026$ & $0.029$ & $0.066$ & $\mathbf{0.008}$ \\ \hline 
    \multirow{3}{*}{\emph{spambase}} & $0.885$ & $0.430$ & $0.575$ & $\mathbf{0.400}$ & $0.415$ & $0.415$ & $0.521$ & $0.620$ & $0.418$ \\ 
          & $\pm$ & $\pm$ & $\pm$ & $\mathbf{\pm}$ & $\pm$ & $\pm$ & $\pm$ & $\pm$ & $\pm$ \\
       & $0.012$ & $0.006$ & $0.010$ & $\mathbf{0.004}$ & $0.004$ & $0.007$ & $0.027$ & $0.043$ & $0.008$ \\ \hline     
    \multirow{3}{*}{\emph{mushroom}} & $0.670$ & $0.383$ & $0.489$ & $0.359$ & $\mathbf{0.350}$ & $0.367$ & $0.502$ & $0.599$ & $0.355$ \\ 
           & $\pm$ & $\pm$ & $\pm$ & $\pm$ & $\mathbf{\pm}$ & $\pm$ & $\pm$ & $\pm$ & $\pm$ \\
       & $0.012$ & $0.006$ & $0.011$ & $0.003$ & $\mathbf{0.006}$ & $0.004$ & $0.032$ & $0.033$ & $0.006$ \\ \hline     
    \end{tabular}
   % \vspace{-3cm}
\end{table}
%\vspace{-1.25cm}

%[TODO: FIGURES HERE]\\

\subsection{Effect of ensemble size}
Next we examined the effect of different ensemble sizes, using $T \in \{10, 25, 50\}$. For this experiment we picked logistic regression as the weak learner. We present results on the stationary datasets.

\begin{table}[H]
\center
\caption{Log-loss across the entire dataset. Lowest average value shown in bold. Results for logistic regression, $T=25$ on stationary datasets. }\label{tab:LR25_stationary}
%\vspace{-0.2cm}
    \begin{tabular}{|c|c|c|c|c|c|c|c|c|c|}
    \hline
 	\multirow{3}{*}{Dataset} &   & Best & Worst &  &  UCB1 &   & Disc. & Disc. & Disc. \\
 	 & Uncalibrated & Fixed & Fixed & UCB1 & Improved & GTS & UCB1 & UCB1 & GTS\\ 
     &  &  &  &  &  &  &  & Improved & \\ \hline\hline    
        \multirow{3}{*}{\emph{landsat}} & $0.323$ & $ 0.120$ & $0.182$ & $0.118$ & $\mathbf{0.113}$ & $0.124$ & $0.202$ & $0.251$ & $0.120$ \\ 
           & $\pm$ & $\pm$ & $\pm$ & $\pm$ & $\pm$ & $\pm$ & $\pm$ & $\pm$ & $\pm$ \\
       & $0.009$ & $0.003$ & $0.001$ & $0.004$ & $\mathbf{0.004}$ & $0.004$ & $0.030$ & $0.029$ & $0.004$ \\ \hline   
    \multirow{3}{*}{\emph{splice}} & $1.553$ & $0.684$ & $0.975$ & $0.607$ & $0.607$ & $0.723$ & $0.667$ & $0.747$ & $\mathbf{0.594}$ \\ 
           & $\pm$ & $\pm$ & $\pm$ & $\pm$ & $\pm$ & $\pm$ & $\pm$ & $\pm$ & $\mathbf{\pm}$ \\
       & $0.064$ & $0.021$ & $0.042$ & $0.016$ & $0.018$ & $0.073$ & $0.058$ & $0.121$ & $\mathbf{0.021}$ \\ \hline    
    \multirow{3}{*}{\emph{musk2}} & $1.515$ & $0.343$ & $0.401$ & $0.343$ & $\mathbf{0.342}$ & $0.344$ & $0.565$ & $0.461$ & $\mathbf{0.342}$ \\ 
          & $\pm$ & $\pm$ & $\pm$ & $\pm$ & $\mathbf{\pm}$ & $\pm$ & $\pm$ & $\pm$ & $\mathbf{\pm}$ \\
       & $0.020$ & $0.003$ & $0.007$ & $0.003$ & $\mathbf{0.002}$ & $0.005$ & $0.037$ & $0.037$ & $\mathbf{0.002}$ \\ \hline 
    \multirow{3}{*}{\emph{krvskp}} & $1.155$ & $0.698$ & $0.895$ & $0.605$ & $\mathbf{0.583}$ & $0.667$ & $0.774$ & $0.765$ & $0.603$ \\ 
           & $\pm$ & $\pm$ & $\pm$ & $\pm$ & $\mathbf{\pm}$ & $\pm$ & $\pm$ & $\pm$ & $\pm$ \\
       & $0.060$ & $0.020$ & $0.040$ & $0.039$ & $\mathbf{0.023}$ & $0.030$ & $0.037$ & $0.042$ & $0.030$ \\ \hline 
    \multirow{3}{*}{\emph{waveform}} & $1.427$ & $0.470$ & $0.572$ & $\mathbf{0.441}$ & $0.457$ & $0.453$ & $0.517$ & $0.551$ & $0.453$ \\ 
          & $\pm$ & $\pm$ & $\pm$ & $\mathbf{\pm}$ & $\pm$ & $\pm$ & $\pm$ & $\pm$ & $\pm$ \\
       & $0.047$ & $0.009$ & $0.011$ & $\mathbf{0.006}$ & $0.019$ & $0.011$ & $0.042$ & $0.048$ & $0.009$ \\ \hline       
    \multirow{3}{*}{\emph{spambase}} & $1.226$ & $0.430$ & $0.590$ & $0.410$ & $\mathbf{0.407}$ & $0.412$ & $0.519$ & $0.524$ & $\mathbf{0.407}$ \\ 
          & $\pm$ & $\pm$ & $\pm$ & $\pm$ & $\mathbf{\pm}$ & $\pm$ & $\pm$ & $\pm$ & $\mathbf{\pm}$ \\
       & $0.019$ & $0.007$ & $0.010$ & $0.006$ & $\mathbf{0.003}$ & $0.005$ & $0.035$ & $0.033$ & $\mathbf{0.007}$ \\ \hline     
    \multirow{3}{*}{\emph{mushroom}} & $1.043$ & $0.362$ & $0.457$ & $\mathbf{0.324}$ & $0.333$ & $0.353$ & $0.431$ & $0.547$ & $0.347$ \\ 
           & $\pm$ & $\pm$ & $\pm$ & $\mathbf{\pm}$ & $\pm$ & $\pm$ & $\pm$ & $\pm$ & $\pm$ \\
       & $0.013$ & $0.005$ & $0.012$ & $\mathbf{0.004}$ & $0.004$ & $0.010$ & $0.031$ & $0.043$ & $0.007$ \\ \hline    
    \end{tabular}
   % \vspace{-3cm}
\end{table}
%\vspace{-1.25cm}

\begin{table}[H]
\center
\caption{Log-loss across the entire dataset. Lowest average value shown in bold. Results for logistic regression, $T=50$ on stationary datasets. }\label{tab:LR50_stationary}
%\vspace{-0.2cm}
    \begin{tabular}{|c|c|c|c|c|c|c|c|c|c|}
    \hline
 	\multirow{3}{*}{Dataset} &   & Best & Worst &  &  UCB1 &   & Disc. & Disc. & Disc. \\
 	 & Uncalibrated & Fixed & Fixed & UCB1 & Improved & GTS & UCB1 & UCB1 & GTS\\ 
     &  &  &  &  &  &  &  & Improved & \\ \hline\hline    
        \multirow{3}{*}{\emph{landsat}} & $0.296$ & $0.132$ & $0.176$ & $0.133$ & $\mathbf{0.122}$ & $0.129$ & $0.151$ & $0.180$ & $0.125$ \\ 
           & $\pm$ & $\pm$ & $\pm$ & $\pm$ & $\mathbf{\pm}$ & $\pm$ & $\pm$ & $\pm$ & $\pm$ \\
       & $0.012$ & $0.005$ & $0.006$ & $0.005$ & $\mathbf{0.005}$ & $0.007$ & $0.017$ & $0.014$ & $0.004$ \\ \hline          
    \multirow{3}{*}{\emph{splice}} & $1.637$ & $0.647$ & $0.907$ & $\mathbf{0.548}$ & $0.577$ & $0.607$ & $0.698$ & $0.652$ & $0.587$ \\ 
           & $\pm$ & $\pm$ & $\pm$ & $\mathbf{\pm}$ & $\pm$ & $\pm$ & $\pm$ & $\pm$ & $\pm$ \\
       & $0.069$ & $0.023$ & $0.050$ & $\mathbf{0.010}$ & $0.026$ & $0.034$ & $0.066$ & $0.030$ & $0.038$ \\ \hline   
    \multirow{3}{*}{\emph{musk2}} & $1.520$ & $0.350$ & $0.403$ & $0.350$ & $\mathbf{0.346}$ & $0.353$ & $0.452$ & $0.550$ & $0.348$ \\ 
           & $\pm$ & $\pm$ & $\pm$ & $\pm$ & $\mathbf{\pm}$ & $\pm$ & $\pm$ & $\pm$ & $\pm$ \\
       & $0.013$ & $0.004$ & $0.008$ & $0.003$ & $\mathbf{0.004}$ & $0.003$ & $0.036$ & $0.045$ & $0.003$ \\ \hline              
    \multirow{3}{*}{\emph{krvskp}} & $1.224$ & $0.675$ & $0.844$ & $\mathbf{0.582}$ & $0.658$ & $0.616$ & $0.701$ & $0.715$ & $0.597$ \\ 
           & $\pm$ & $\pm$ & $\pm$ & $\mathbf{\pm}$ & $\pm$ & $\pm$ & $\pm$ & $\pm$ & $\pm$ \\
       & $0.045$ & $0.024$ & $0.032$ & $\mathbf{0.010}$ & $0.033$ & $0.029$ & $0.024$ & $0.020$ & $0.009$ \\ \hline 
    \multirow{3}{*}{\emph{waveform}} & $1.439$ & $0.470$ & $0.591$ & $0.482$ & $0.444$ & $0.482$ & $0.556$ & $0.563$ & $\mathbf{0.441}$ \\ 
           & $\pm$ & $\pm$ & $\pm$ & $\pm$ & $\pm$ & $\pm$ & $\pm$ & $\pm$ & $\mathbf{\pm}$ \\
       & $0.033$ & $0.005$ & $0.015$ & $0.031$ & $0.005$ & $0.028$ & $0.030$ & $0.027$ & $\mathbf{0.005}$ \\ \hline 
    \multirow{3}{*}{\emph{spambase}} & $1.241$ & $0.429$ & $0.584$ & $\mathbf{0.397}$ & $0.399$ & $0.410$ & $0.527$ & $0.527$ & $0.414$ \\ 
           & $\pm$ & $\pm$ & $\pm$ & $\mathbf{\pm}$ & $\pm$ & $\pm$ & $\pm$ & $\pm$ & $\pm$ \\
       & $0.018$ & $0.005$ & $0.011$ & $\mathbf{0.006}$ & $0.006$ & $0.007$ & $0.028$ & $0.032$ & $0.005$ \\ \hline          
    \multirow{3}{*}{\emph{mushroom}} & $1.082$ & $0.363$ & $0.481$ & $0.345$ & $0.344$ & $\mathbf{0.340}$ & $0.464$ & $0.529$ & $0.345$ \\ 
           & $\pm$ & $\pm$ & $\pm$ & $\pm$ & $\pm$ & $\mathbf{\pm}$ & $\pm$ & $\pm$ & $\pm$ \\
       & $0.017$ & $0.004$ & $0.010$ & $0.005$ & $0.006$ & $\mathbf{0.009}$ & $0.040$ & $0.044$ & $0.008$ \\ \hline    
    \end{tabular}
   % \vspace{-3cm}
\end{table}
%\vspace{-1.25cm}

\subsection{Effect of weak learner regularization}
We now investigate different degrees of regularization on the weak learner. We used $\ell_1$-regularized logistic regression with a regularization parameter $\lambda \in \{10^{-1}, 10^{-2}, 10^{-3},0 \}$, with $T=10$. We present experiments on the stationary datasets.

\begin{table}[H]
\center

\caption{Log-loss across the entire dataset. Lowest average value shown in bold. Results for $\ell_1$-regularized logistic regression, $T=10$ and $\lambda=10^{-3}$ on stationary datasets. }\label{tab:LR10_stationary_lambda_0_001}
%\vspace{-0.2cm}
    \begin{tabular}{|c|c|c|c|c|c|c|c|c|c|}
    \hline
 	\multirow{3}{*}{Dataset} &   & Best & Worst &  &  UCB1 &   & Disc. & Disc. & Disc. \\
 	 & Uncalibrated & Fixed & Fixed & UCB1 & Improved & GTS & UCB1 & UCB1 & GTS\\ 
     &  &  &  &  &  &  &  & Improved & \\ \hline\hline    
        \multirow{3}{*}{\emph{landsat}} & $0.335$ & $0.105$ & $0.153$ & $\mathbf{0.101}$ & $0.103$ & $0.105$ & $0.150$ & $0.172$ & $0.118$ \\ 
           & $\pm$ & $\pm$ & $\pm$ & $\mathbf{\pm}$ & $\pm$ & $\pm$ & $\pm$ & $\pm$ & $\pm$ \\
       & $0.006$ & $0.003$ & $0.004$ & $\mathbf{0.003}$ & $0.002$ & $0.003$ & $0.015$ & $0.027$ & $0.005$ \\ \hline
    \multirow{3}{*}{\emph{splice}} & $1.761$ & $0.604$ & $0.885$ & $0.520$ & $0.576$ & $\mathbf{0.502}$ & $0.720$ & $0.707$ & $0.517$ \\ 
           & $\pm$ & $\pm$ & $\pm$ & $\pm$ & $\pm$ & $\mathbf{\pm}$ & $\pm$ & $\pm$ & $\pm$ \\
       & $0.093$ & $0.024$ & $0.040$ & $0.022$ & $0.070$ & $\mathbf{0.013}$ & $0.135$ & $0.076$ & $0.019$ \\ \hline   
    \multirow{3}{*}{\emph{musk2}} & $1.334$ & $0.335$ & $0.424$ & $0.320$ & $\mathbf{0.319}$ & $0.317$ & $0.480$ & $0.683$ & $0.324$ \\ 
           & $\pm$ & $\pm$ & $\pm$ & $\pm$ & $\mathbf{\pm}$ & $\pm$ & $\pm$ & $\pm$ & $\pm$ \\
       & $0.022$ & $0.006$ & $0.008$ & $0.004$ & $\mathbf{0.004}$ & $0.004$ & $0.048$ & $0.122$ & $0.003$ \\ \hline       
    \multirow{3}{*}{\emph{krvskp}} & $1.126$ & $0.592$ & $0.805$ & $\mathbf{0.479}$ & $0.605$ & $0.546$ & $0.781$ & $0.823$ & $0.581$ \\ 
           & $\pm$ & $\pm$ & $\pm$ & $\mathbf{\pm}$ & $\pm$ & $\pm$ & $\pm$ & $\pm$ & $\pm$ \\
       & $0.036$ & $0.007$ & $0.029$ & $\mathbf{0.016}$ & $0.051$ & $0.039$ & $0.075$ & $0.077$ & $0.055$ \\ \hline
    \multirow{3}{*}{\emph{waveform}} & $1.879$ & $0.457$ & $0.609$ & $0.440$ & $0.440$ & $\mathbf{0.437}$ & $0.523$ & $0.642$ & $0.459$ \\ 
           & $\pm$ & $\pm$ & $\pm$ & $\pm$ & $\pm$ & $\mathbf{\pm}$ & $\pm$ & $\pm$ & $\pm$ \\
       & $0.044$ & $0.007$ & $0.012$ & $0.007$ & $0.006$ & $\mathbf{0.007}$ & $0.032$ & $0.135$ & $0.011$ \\ \hline 
    \multirow{3}{*}{\emph{spambase}} & $0.994$ & $0.406$ & $0.567$ & $\mathbf{0.373}$ & $0.375$ & $0.383$ & $0.506$ & $0.556$ & $ 0.386$ \\ 
           & $\pm$ & $\pm$ & $\pm$ & $\mathbf{\pm}$ & $\pm$ & $\pm$ & $\pm$ & $\pm$ & $\pm$ \\
       & $0.024$ & $0.006$ & $0.005$ & $\mathbf{0.006}$ & $0.006$ & $0.005$ & $0.024$ & $0.034$ & $0.009$ \\ \hline
    \multirow{3}{*}{\emph{mushroom}} & $0.891$ & $0.317$ & $0.461$ & $\mathbf{0.305}$ & $0.312$ & $0.314$ & $0.462$ & $0.590$ & $0.319$ \\ 
           & $\pm$ & $\pm$ & $\pm$ & $\mathbf{\pm}$ & $\pm$ & $\pm$ & $\pm$ & $\pm$ & $\pm$ \\
       & $0.009$ & $0.005$ & $0.014$ & $\mathbf{0.005}$ & $0.006$ & $0.008$ & $0.045$ & $0.033$ & $0.006$ \\ \hline    
    \end{tabular}
   % \vspace{-3cm}
\end{table}
%\vspace{-1.25cm}

\begin{table}[htb]
\center
\caption{Log-loss across the entire dataset. Lowest average value shown in bold. Results for $\ell_1$-regularized logistic regression, $T=10$ and $\lambda=10^{-2}$ on stationary datasets. }\label{tab:LR10_stationary_lambda_0_01}
%\vspace{-0.2cm}
    \begin{tabular}{|c|c|c|c|c|c|c|c|c|c|}
    \hline
 	\multirow{3}{*}{Dataset} &   & Best & Worst &  &  UCB1 &   & Disc. & Disc. & Disc. \\
 	 & Uncalibrated & Fixed & Fixed & UCB1 & Improved & GTS & UCB1 & UCB1 & GTS\\ 
     &  &  &  &  &  &  &  & Improved & \\ \hline\hline    
        \multirow{3}{*}{\emph{landsat}} & $0.345$ & $0.110$ & $0.157$ & $0.103$ & $0.106$ & $0.107$ & $0.144$ & $0.147$ & $\mathbf{0.102}$ \\ 
           & $\pm$ & $\pm$ & $\pm$ & $\pm$ & $\pm$ & $\pm$ & $\pm$ & $\pm$ & $\mathbf{\pm}$ \\
       & $0.009$ & $0.003$ & $0.004$ & $0.003$ & $0.005$ & $0.002$ & $0.012$ & $0.017$ & $\mathbf{0.003}$ \\ \hline            
    \multirow{3}{*}{\emph{splice}} & $1.483$ & $0.572$ & $0.806$ & $0.529$ & $\mathbf{0.489}$ & $0.528$ & $0.556$ & $0.704$ & $0.517$ \\ 
           & $\pm$ & $\pm$ & $\pm$ & $\pm$ & $\mathbf{\pm}$ & $\pm$ & $\pm$ & $\pm$ & $\pm$ \\
       & $0.054$ & $0.025$ & $0.033$ & $0.015$ & $\mathbf{0.011}$ & $0.041$ & $0.022$ & $0.076$ & $0.013$ \\ \hline   
    \multirow{3}{*}{\emph{musk2}} & $1.501$ & $0.352$ & $0.455$ & $0.348$ & $0.349$ & $\mathbf{0.347}$ & $0.481$ & $0.693$ & $0.360$ \\ 
           & $\pm$ & $\pm$ & $\pm$ & $\pm$ & $\pm$ & $\mathbf{\pm}$ & $\pm$ & $\pm$ & $\pm$ \\
       & $0.016$ & $0.004$ & $0.003$ & $0.004$ & $0.004$ & $\mathbf{0.002}$ & $0.038$ & $0.145$ & $0.006$ \\ \hline 
    \multirow{3}{*}{\emph{krvskp}} & $1.357$ & $0.756$ & $1.007$ & $0.658$ & $0.628$ & $0.654$ & $0.782$ & $0.776$ & $\mathbf{0.583}$ \\ 
           & $\pm$ & $\pm$ & $\pm$ & $\pm$ & $\pm$ & $\pm$ & $\pm$ & $\pm$ & $\mathbf{\pm}$ \\
       & $0.071$ & $0.05$ & $0.05$ & $0.039$ & $0.027$ & $0.029$ & $0.064$ & $0.07$ & $\mathbf{0.023}$ \\ \hline
    \multirow{3}{*}{\emph{waveform}} & $1.491$ & $0.456$ & $0.629$ & $\mathbf{0.434}$ & $\mathbf{0.434}$ & $0.444$ & $0.494$ & $0.515$ & $0.441$ \\ 
           & $\pm$ & $\pm$ & $\pm$ & $\pm$ & $\pm$ & $\pm$ & $\pm$ & $\pm$ & $\pm$ \\
       & $0.031$ & $0.006$ & $0.020$ & $\mathbf{0.006}$ & $\mathbf{0.006}$ & $0.006$ & $0.024$ & $0.035$ & $0.007$ \\ \hline 
    \multirow{3}{*}{\emph{spambase}} & $0.887$ & $0.405$ & $0.553$ & $0.383$ & $0.380$ & $0.400$ & $0.475$ & $0.556$ & $\mathbf{0.379}$ \\ 
           & $\pm$ & $\pm$ & $\pm$ & $\pm$ & $\pm$ & $\pm$ & $\pm$ & $\pm$ & $\mathbf{\pm}$ \\
       & $0.026$ & $0.007$ & $0.007$ & $0.004$ & $0.006$ & $0.009$ & $0.024$ & $0.063$ & $\mathbf{0.004}$ \\ \hline         
    \multirow{3}{*}{\emph{mushroom}} & $0.743$ & $0.330$ & $0.465$ & $\mathbf{0.310}$ & $\mathbf{0.310}$ & $0.318$ & $0.790$ & $0.495$ & $0.320$ \\ 
           & $\pm$ & $\pm$ & $\pm$ & $\mathbf{\pm}$ & $\mathbf{\pm}$ & $\pm$ & $\pm$ & $\pm$ & $\pm$ \\
       & $0.012$ & $0.003$ & $0.011$ & $\mathbf{0.005}$ & $\mathbf{0.004}$ & $0.007$ & $0.031$ & $0.037$ & $0.012$ \\ \hline    
    \end{tabular}
   % \vspace{-3cm}
\end{table}
%\vspace{-1.25cm}		

\begin{table}[htb]
\center
\caption{Log-loss across the entire dataset. Lowest average value shown in bold. Results for $\ell_1$-regularized logistic regression, $T=10$ and $\lambda=10^{-1}$ on stationary datasets. }\label{tab:LR10_stationary_lambda_0_1}
%\vspace{-0.2cm}
    \begin{tabular}{|c|c|c|c|c|c|c|c|c|c|}
    \hline
 	\multirow{3}{*}{Dataset} &   & Best & Worst &  &  UCB1 &   & Disc. & Disc. & Disc. \\
 	 & Uncalibrated & Fixed & Fixed & UCB1 & Improved & GTS & UCB1 & UCB1 & GTS\\ 
     &  &  &  &  &  &  &  & Improved & \\ \hline\hline    
        \multirow{3}{*}{\emph{landsat}} & $0.308$ & $0.113$ & $0.153$ & $0.100$ & $0.100$ & $\mathbf{0.099}$ & $0.141$ & $0.142$ & $0.106$ \\ 
           & $\pm$ & $\pm$ & $\pm$ & $\pm$ & $\pm$ & $\mathbf{\pm}$ & $\pm$ & $\pm$ & $\pm$ \\
       & $0.010$ & $0.002$ & $0.006$ & $0.003$ & $0.002$ & $\mathbf{0.002}$ & $0.013$ & $0.017$ & $0.004$ \\ \hline 
    \multirow{3}{*}{\emph{splice}} & $3.703$ & $0.930$ & $1.607$ & $0.778$ & $\mathbf{0.770}$ & $0.927$ & $1.187$ & $0.988$ & $0.883$ \\ 
           & $\pm$ & $\pm$ & $\pm$ & $\pm$ & $\mathbf{\pm}$ & $\pm$ & $\pm$ & $\pm$ & $\pm$ \\
       & $0.119$ & $0.037$ & $0.080$ & $0.022$ & $\mathbf{0.023}$ & $0.091$ & $0.288$ & $0.267$ & $0.098$ \\ \hline  
    \multirow{3}{*}{\emph{musk2}} & $2.244$ & $0.510$ & $0.794$ & $\mathbf{0.462}$ & $0.466$ & $0.503$ & $0.664$ & $0.492$ & $0.466$ \\ 
           & $\pm$ & $\pm$ & $\pm$ & $\pm$ & $\mathbf{\pm}$ & $\pm$ & $\pm$ & $\pm$ & $\pm$ \\
       & $0.038$ & $0.006$ & $0.012$ & $\mathbf{0.008}$ & $0.007$ & $0.035$ & $0.161$ & $0.009$ & $0.013$ \\ \hline 
    \multirow{3}{*}{\emph{krvskp}} & $2.857$ & $1.109$ & $1.483$ & $0.775$ & $0.797$ & $\mathbf{0.749}$ & $0.924$ & $0.923$ & $0.786$ \\ 
           & $\pm$ & $\pm$ & $\pm$ & $\pm$ & $\pm$ & $\mathbf{\pm}$ & $\pm$ & $\pm$ & $\pm$ \\
       & $0.263$ & $0.044$ & $0.084$ & $0.053$ & $0.037$ & $\mathbf{0.030}$ & $0.143$ & $0.149$ & $0.030$ \\ \hline 
    \multirow{3}{*}{\emph{waveform}} & $1.198$ & $0.557$ & $0.698$ & $0.500$ & $0.552$ & $0.488$ & $0.569$ & $0.639$ & $\mathbf{0.486}$ \\ 
           & $\pm$ & $\pm$ & $\pm$ & $\pm$ & $\pm$ & $\pm$ & $\pm$ & $\pm$ & $\mathbf{\pm}$ \\
       & $0.050$ & $0.014$ & $0.025$ & $0.011$ & $0.030$ & $0.005$ & $0.023$ & $0.030$ & $\mathbf{0.007}$ \\ \hline 
    \multirow{3}{*}{\emph{spambase}} & $0.639$ & $0.563$ & $0.671$ & $\mathbf{0.491}$ & $0.516$ & $0.511$ & $0.597$ & $0.593$ & $0.515$ \\ 
           & $\pm$ & $\pm$ & $\pm$ & $\mathbf{\pm}$ & $\pm$ & $\pm$ & $\pm$ & $\pm$ & $\pm$ \\
       & $0.014$ & $0.007$ & $0.015$ & $\mathbf{0.008}$ & $0.008$ & $0.011$ & $0.020$ & $0.015$ & $0.008 $ \\ \hline    
    \multirow{3}{*}{\emph{mushroom}} & $0.521$ & $0.428$ & $0.564$ & $0.394$ & $\mathbf{0.378}$ & $0.404$ & $0.553$ & $0.628$ & $0.400$ \\ 
           & $\pm$ & $\pm$ & $\pm$ & $\pm$ & $\mathbf{\pm}$ & $\pm$ & $\pm$ & $\pm$ & $\pm$ \\
       & $0.009$ & $0.005$ & $0.007$ & $0.007$ & $\mathbf{0.005}$ & $0.007$ & $0.018$ & $0.021$ & $0.009$ \\ \hline    
    \end{tabular}
   % \vspace{-3cm}
\end{table}
%\vspace{-1.25cm}

%[TODO: FIGURES HERE]\\

\end{document}